\PassOptionsToPackage{table}{xcolor}
\documentclass{article}

\PassOptionsToPackage{numbers, compress}{natbib}
\usepackage[main, final]{neurips_2026}

\usepackage[utf8]{inputenc} % allow utf-8 input
\usepackage[T1]{fontenc}    % use 8-bit T1 fonts
\usepackage{hyperref}   
\usepackage{booktabs}       % professional-quality tables
\usepackage{amsfonts}       % blackboard math symbols
\usepackage{amsmath}        % math environments and \text
\usepackage{nicefrac}       % compact symbols for 1/2, etc.
\usepackage{microtype}      % microtypography
\usepackage{xcolor}
\usepackage{makecell}       % multi-line cells
\usepackage{multirow}       % multi-row cells
\usepackage{wrapfig}        % wrap text around figures
\usepackage{graphicx}
\usepackage{arydshln}       % dashed lines in tables
\usepackage{algorithm}
\usepackage{algorithmic}
\usepackage{CJKutf8}        % Chinese character support

\newenvironment{compactmath}{%
  \begingroup
  \small
  \setlength{\abovedisplayskip}{4pt}%
  \setlength{\belowdisplayskip}{4pt}%
  \setlength{\abovedisplayshortskip}{2pt}%
  \setlength{\belowdisplayshortskip}{2pt}%
  \setlength{\jot}{2pt}%
}{\endgroup}

\newif\ifshowcomments
\showcommentstrue             % set to \showcommentsfalse to hide all comments

\title{Beyond SFT-to-RL: Pre-alignment via Black-box \\ On-policy  Distillation for Multimodal RL}

\author{%
  \textbf{Sudong Wang}$^{1*}$\quad
  \textbf{Weiquan Huang}$^{1*}$\quad
  \textbf{Xiaomin Yu}$^{1}$\quad
  \textbf{Zuhao Yang}$^{3}$\quad
  \textbf{Hehai Lin}$^{1}$\\
  \textbf{Keming Wu}$^{2}$\quad
  \textbf{Chaojun Xiao}$^{2}$\quad
  \textbf{Chen Chen}$^{3}$\quad
  \textbf{Wenxuan Wang}$^{4}$\quad
  \textbf{Beier Zhu}$^{5}$\\
  \textbf{Yunjian Zhang}$^{6\dagger}$\quad
  \textbf{Chengwei Qin}$^{1,7\dagger}$\\[4pt]
  $^{1}$Hong Kong University of Science and Technology (Guangzhou)\\
  $^{2}$Tsinghua University\quad
  $^{3}$Nanyang Technological University\\
  $^{4}$Renmin University of China\quad
  $^{5}$University of Science and Technology of China\\
  $^{6}$University of Chinese Academy of Sciences\\
  $^{7}$Hong Kong University of Science and Technology\\[2pt]
  \texttt{\{swang886, whuang491\}@connect.hkust-gz.edu.cn}
}

\usepackage{tcolorbox}
\usepackage{enumitem}
\tcbuselibrary{skins, breakable}

\definecolor{blueboxtitle}{HTML}{2C4A5C}
\definecolor{blueboxbg}{HTML}{D6EAF2}
\definecolor{blueboxframe}{HTML}{8ABBD4}

\definecolor{greenboxtitle}{HTML}{2E7D32}
\definecolor{greenboxbg}{HTML}{E1F0E1}
\definecolor{greenboxframe}{HTML}{81C784}

\begin{document}

\maketitle
\renewcommand{\thefootnote}{\fnsymbol{footnote}}
\footnotetext[1]{Equal contribution.}
\footnotetext[2]{Corresponding authors.}
\renewcommand{\thefootnote}{\arabic{footnote}}

\begin{abstract}
The standard post-training recipe for large multimodal models (LMMs) applies supervised fine-tuning (SFT) on curated demonstrations followed by reinforcement learning with verifiable rewards (RLVR). However, SFT introduces distributional drift that neither preserves the model's original capabilities nor faithfully matches the supervision distribution. This problem is further amplified in multimodal reasoning, where perception errors and reasoning failures follow distinct drift patterns that compound during subsequent RL.
We introduce \textbf{PRISM}, a three-stage pipeline that mitigates this drift by inserting an explicit distribution-alignment stage between SFT and RLVR.
Building on the principle of on-policy distillation (OPD), PRISM casts alignment as a black-box, response-level adversarial game between the policy and a Mixture-of-Experts (MoE) discriminator with dedicated perception and reasoning experts, providing disentangled corrective signals that steer the policy toward the supervision distribution without requiring access to teacher logits.
While 1.26M public demonstrations suffice for broad SFT initialization, distribution alignment demands higher-fidelity supervision; we therefore curate 113K additional demonstrations from Gemini~3 Flash, featuring dense visual grounding and step-by-step reasoning on the hardest unsolved problems.
Experiments on Qwen3-VL show that PRISM consistently improves downstream RLVR performance across multiple RL algorithms (GRPO, DAPO, GSPO) and diverse multimodal benchmarks, improving average accuracy by +4.4 and +6.0 points over the SFT$\rightarrow$RLVR baseline on 4B and 8B, respectively. Our code, data, and model checkpoints are publicly available at \href{https://github.com/XIAO4579/PRISM}{\texttt{https://github.com/XIAO4579/PRISM}}.
\end{abstract}

\section{Introduction}
\label{sec:intro}
Driven by the success of reasoning-oriented large language models (LLMs)~\citep{guo2025deepseek,jaech2024openai,yang2025qwen3,zeng2026glm,lin2026unified}, large multimodal models (LMMs) have also demonstrated strong instruction-following and reasoning capabilities~\citep{bai2025qwen3,chen2024expanding}. A prevailing paradigm for improving such capabilities is a two-stage post-training pipeline: models are first adapted via offline supervised fine-tuning (SFT) on curated demonstrations \citep{liu2023visual,liu2024improved}, and then further optimized with online reinforcement learning with verifiable rewards (RLVR) \citep{shao2024deepseekmath,yang2024qwen2}, which directly improves task performance using automatic verifiers. In this pipeline, SFT provides a crucial capability bootstrap by anchoring the model to high-quality supervision, while RLVR further refines the policy toward task-specific objectives and largely determines the final performance. As a result, a growing body of work has focused on improving the effectiveness and stability of both stages. For SFT, recent methods optimize it by reweighting or regularizing next-token likelihood \citep{qin2025supervised,zhu2025proximal}. For RLVR, a number of approaches have been proposed to improve optimization stability and reduce variance, including GRPO-style variants that redesign importance weighting and clipping mechanisms to stabilize policy updates \citep{yu2025dapo,zheng2025group,zhao2025geometric,yue2025vapo,wang2026calibration}. The underlying intuition is straightforward: SFT establishes an implicit reasoning prior in the model's parameter space, whereas RLVR activates and refines this capability through online optimization \citep{chu2025sft,yue2025does}. 

However, recent studies have uncovered a striking and counterintuitive phenomenon: instead of reliably improving the model, offline supervision may place the model in a compromised state, where it neither adequately matches the demonstration policy distribution nor retains the model's original favorable distribution \citep{kang2025quagmires,zhang2026good}. In this sense, SFT can become a source of distributional drift rather than a pure improvement step. %Our further experiments indicate that this phenomenon is especially evident in models with stronger priors. 
A plausible explanation is that SFT optimizes the model to imitate trajectories sampled from the demonstration policy under a uniform token-level objective, without distinguishing between process and outcome. 
As a result, the model may learn surface-level patterns rather than faithful reasoning capabilities, and simultaneously drift away from its original distribution. While this drift is often tolerable for weaker models that gain substantially from learning the demonstration policy, it becomes increasingly costly as the base model grows stronger: when the model already possesses a capable reasoning distribution, token-level imitation of an external demonstration policy can displace the model's native strengths rather than supplement them~\citep{zhang2026good,kang2025quagmires}. This issue becomes particularly pronounced in multimodal models, where the distributional bias introduced by SFT interacts with imperfect visual grounding: even slight deviations at the perception stage can distort the premises of reasoning and subsequently amplify errors throughout RL \citep{liu2025more,chu2025sft}.
Moreover, unlike the relatively uniform drift in text-only models, multimodal drift is inherently heterogeneous: visual grounding and logical reasoning degrade in qualitatively different ways that a single corrective objective cannot jointly address. This raises a natural question: \textbf{How can we repair the distributional drift introduced by SFT, particularly its heterogeneous impact on visual perception and reasoning, before the model enters RL?}

\begin{figure}[t]
    \centering
    \includegraphics[width=\textwidth]{figures/overview.pdf}
    \vspace{-4mm}
    \caption{Overview of the PRISM pipeline. (a) SFT introduces distributional drift between the policy and the supervision distribution. (b) The alignment stage uses an MoE discriminator with dedicated perception and reasoning experts to repair this drift via adversarial on-policy distillation. (c) The resulting distribution-aligned policy provides a stronger initialization for downstream RLVR.}
    \label{fig:overview}
     \vspace{-6mm}
\end{figure}

Advances in knowledge distillation suggest that a model can benefit substantially from learning from its own on-policy generations rather than relying solely on static teacher-forced targets~\citep{gu2024minillm,agarwal2024policy,zhao2026self}. By optimizing on rollouts sampled from its current policy, on-policy distillation (OPD) mitigates exposure bias and encourages more faithful policy refinement \citep{gu2024minillm,zhang2019bridging}. Building on this principle, we propose \textbf{PR}e-alignment via black-box on-policy d\textbf{IS}tillation for \textbf{M}ultimodal reinforcement learning (PRISM), a new three-stage post-training paradigm that extends the standard SFT$\rightarrow$RL recipe with an explicit pre-alignment stage. The core of PRISM is an adversarial OPD framework that drives the post-SFT policy distribution toward the supervision distribution, while introducing a logit-free formulation that eliminates the external-teacher dependency of standard OPD. Specifically, we formulate alignment as a minimax game \citep{goodfellow2020generative} between the policy and a Mixture-of-Experts (MoE) discriminator with dedicated vision and reasoning experts. The discriminator learns to separate policy rollouts from the supervision pool by probing both perceptual grounding and reasoning consistency, while the policy is optimized to generate responses that increasingly resemble the supervision distribution. This design establishes a critical distribution-level alignment stage after SFT, not only correcting distributional drift but also preparing a more reliable initialization for online optimization.

We validate PRISM on Qwen3-VL across diverse multimodal benchmarks and multiple RL algorithms. The results confirm consistent improvements over the standard SFT$\rightarrow$RLVR pipeline, and further analysis shows that the alignment stage substantially narrows the distributional gap left by SFT. An overview of the PRISM pipeline is shown in Figure~\ref{fig:overview}.

In summary, our main contributions are as follows:
\begin{itemize}[topsep=2pt,itemsep=2pt,parsep=0pt,leftmargin=*]
    \item We identify the distributional mismatch left by SFT as a bottleneck for multimodal RLVR, and propose PRISM, which addresses it through black-box adversarial on-policy pre-alignment before RLVR, using an MoE discriminator that provides decoupled corrective signals for perception and reasoning drift.
    \item We curate a 113K high-quality multimodal reasoning corpus distilled from Gemini 3 Flash, targeting the hardest problems unsolved by current LMMs with dense visual grounding and step-by-step reasoning traces. Combined with 1.26M publicly available demonstrations from the same model family, this corpus serves as both the SFT foundation and the supervision reference for distribution alignment.
    \item Experiments on Qwen3-VL-4B/8B validate that PRISM consistently and substantially improves downstream RLVR, with PRISM+GRPO outperforming SFT$\rightarrow$GRPO by +4.4 and +6.0 average points on the two scales, respectively, and similar gains are observed across DAPO and GSPO.
\end{itemize}

\section{Related Work}

\subsection{Reinforcement Learning for Multimodal Reasoning}
Reinforcement learning with verifiable rewards (RLVR) has emerged as a dominant paradigm for improving reasoning in both large language models and large multimodal models (LMMs). In the text domain, DeepSeek-R1~\citep{guo2025deepseek} demonstrated that pure RL with verifiable rewards can elicit emergent chain-of-thought reasoning without human-labeled traces, motivating a series of algorithmic improvements that enhance optimization stability at scale through redesigned clipping, advantage estimation, critic-free architectures, or sequence-level objectives~\citep{shao2024deepseekmath,hu2025reinforce++,yu2025dapo,liu2025understanding,zheng2025group}. In the multimodal domain, early efforts explored R1-style RL for LMMs via cold-start initialization~\citep{huang2025vision}, cross-modal formalization~\citep{yang2025r1}, large-scale rule-based RL with emergent reflection~\citep{meng2025mm,zhang2025openmmreasoner}, curriculum-based sampling~\citep{hong2025glm}, self-reflection incentivization~\citep{wang2025vl}, and tool-augmented agentic video reasoning~\citep{yang2026paravt}. More recently, a line of work has recognized that vanilla RLVR neglects visual perception fidelity, and proposed perception-aware reward signals through judging LLMs~\citep{xiao2025perception}, evidence-anchored dual-branch reasoning~\citep{zhang2025perceptual}, or differential visual reasoning with visual triplets~\citep{gao2026thinking}. While these methods advance multimodal reasoning through better RL algorithms or reward designs, their improvements all take place $\textit{within}$ the RL stage, leaving untouched the distributional drift introduced by SFT, which compromises the starting point of RL optimization. PRISM does not improve RL itself, but instead inserts black-box on-policy distillation as a bridging stage between SFT and RL to correct this drift.

\subsection{On-Policy Distillation}
Standard knowledge distillation for LLMs performs SFT on teacher-generated outputs, but this off-policy approach suffers from a distribution mismatch between training and inference. On-policy distillation (OPD) addresses this by training the student on its own generations: GKD~\citep{agarwal2024policy} introduced the paradigm with flexible divergence objectives, followed by explorations of alternative divergences~\citep{gu2024minillm,ko2024distillm} and logit-free adversarial formulations~\citep{ye2025black}. Recent extensions further broaden OPD along complementary axes such as self-distillation~\citep{zhao2026self}, reward extrapolation~\citep{yang2026learning}, selective imitation~\citep{zhang2026reinforcement}, and multimodal representation transfer~\citep{cai2025llava}. Despite these advances, most existing OPD methods treat distillation as a terminal training objective where the resulting checkpoint serves directly as the final model, and rely on a single undifferentiated discriminator or divergence signal. PRISM instead positions OPD as an intermediate alignment stage that explicitly prepares the policy for subsequent RLVR, and employs an MoE discriminator with dedicated vision and reasoning experts to provide decoupled rewards that address the heterogeneous nature of multimodal distribution shift. In the multimodal setting, VOLD~\citep{bousselham2025vold} combines GRPO with logit-based on-policy distillation from a text-only teacher into a unified training objective. PRISM differs in three key respects: it decouples alignment from RL as a standalone intermediate stage, it operates without teacher logits via adversarial discrimination, and it provides decoupled feedback through dedicated perception and reasoning experts.

\section{Method}
\label{sec:method}

We present PRISM, a three-stage post-training pipeline that augments the conventional SFT$\rightarrow$RL paradigm with an intermediate pre-alignment stage. Specifically, PRISM first performs SFT on high-quality demonstrations to obtain an initial policy, then applies adversarial OPD with an MoE discriminator to recalibrate the post-SFT policy distribution, and finally conducts outcome-based RLVR for final policy improvement. The complete procedure is provided in Appendix~\ref{app:algorithm}; we describe each stage in turn below.

\subsection{Cold-Start Supervised Fine-Tuning}
\label{sec:sft}
As the first stage of PRISM, SFT serves as a cold start that equips the model with an initial multimodal reasoning policy. Since the same supervision source is later reused in the alignment stage for distribution-level correction, each sample must contain not only a correct final answer but also a complete reasoning trajectory with accurate visual grounding. Existing public multimodal datasets are often inadequate for this purpose, as many contain brief answers, incomplete reasoning traces, or imprecise visual descriptions.

To address this, we curate a 113K multimodal reasoning corpus following~\citep{ye2025limo,lin2026mmfinereason}: we collect problems with zero pass rate under strong contemporary models, generate detailed solutions with Gemini 3 Flash~\citep{gemini3flash_website_2025} requiring fine-grained visual grounding and step-by-step deduction, and apply multi-stage filtering including format validation and LLM-based correctness verification (details in Appendix~\ref{app:data}). Among the resulting samples, 107K are used for SFT and the remaining 6K are reserved for the alignment and RL stages. Since a policy trained on insufficient data remains far from the target distribution, placing an excessive corrective burden on the downstream alignment stage, we supplement our curated corpus with 1.26M publicly available demonstrations from the same Gemini model family~\citep{leng2025mmr1}, yielding a combined SFT corpus of approximately 1.37M samples.

Using the combined corpus, we perform standard supervised fine-tuning to obtain an initial reasoning-capable policy. As we show next, however, SFT alone does not guarantee a distribution well suited for subsequent RL optimization, motivating the explicit pre-alignment stage.

\subsection{Distribution Alignment via On-Policy Distillation}
\label{sec:alignment}

\subsubsection{Overview}
The alignment stage repairs the distributional drift introduced by SFT before the model enters RLVR. As discussed in Section~\ref{sec:intro}, the post-SFT policy may only partially absorb the target behavior while drifting away from its native distribution. Directly passing such a policy to RLVR forces online optimization to start from a distorted state, limiting the gains that RL can deliver. A natural idea is to apply additional token-level imitation, but this only encourages surface-level matching without repairing the mismatch that emerges under on-policy generation. Moreover, the supervision data may originate from proprietary black-box models whose logits are inaccessible, rendering divergence-based distillation inapplicable. We therefore formulate alignment as a response-level adversarial game that requires only samples from the supervision pool. A further challenge is that distributional drift in multimodal reasoning is inherently heterogeneous: visual grounding errors and reasoning failures require qualitatively different corrections. This motivates a Mixture-of-Experts discriminator with dedicated perception and reasoning experts. The overall architecture is illustrated in Figure~\ref{fig:pipeline}.

\begin{figure}[t]
    \centering
    \includegraphics[width=0.9\textwidth]{figures/alignment_pipeline.pdf}
    \caption{Architecture of the distribution-alignment stage. An MoE discriminator with perception and reasoning experts is trained via Bradley-Terry loss to distinguish supervision from policy outputs; the policy is updated via policy gradient to maximize the combined MoE reward.}
    \label{fig:pipeline}
    \vspace{-4mm}
\end{figure}

\subsubsection{Mixture-of-Experts Discriminator}

To provide targeted corrective signals for heterogeneous errors in multimodal reasoning, we instantiate the alignment module with an MoE discriminator. The key idea is that deviations from the supervision distribution typically arise from two distinct sources: failures in visual grounding and failures in logical reasoning. A single discriminator is often too coarse to capture these two error modes simultaneously. We therefore decompose the discrimination process into two specialized experts, each responsible for one aspect of the response. 
Concretely, each response $y$ for multimodal input $x$ consists of a visual description $c$ and a reasoning trace $t$. We define two experts in the discriminator:
\begin{itemize}[topsep=2pt,itemsep=2pt,parsep=0pt,leftmargin=*]
    \item \textbf{Perception Expert $D_v$}: evaluates the visual description $c$ and measures how well the response is grounded in the visual input;
    \item \textbf{Reasoning Expert $D_r$}: evaluates the reasoning trace $t$ and measures the consistency and validity of the underlying deduction.
\end{itemize}
The discriminator score is then defined as a weighted combination of the two expert scores:
\begin{compactmath}
\begin{equation}
\label{eq:moe_reward}
r(x, y) = \alpha \cdot D_v(x, c) + (1 - \alpha) \cdot D_r(x, t),
\end{equation}
\end{compactmath}
where $\alpha$ controls the trade-off between perceptual and reasoning feedback.

By delivering disentangled feedback on the two dominant sources of multimodal error rather than collapsing them into a single scalar, the MoE discriminator provides a finer-grained basis for the adversarial alignment objective introduced next.

\subsubsection{Initialization for Alignment}

The adversarial alignment stage assumes that the policy and the discriminator are reasonably matched in capability. In practice, however, a pretrained LMM before alignment remains far from the supervision distribution, making its responses trivially separable from reference demonstrations. Under such a large gap, the discriminator can quickly saturate, leaving the policy with uninformative training signals~\citep{goodfellow2014generative,arjovsky2017wasserstein}. We therefore initialize both components before entering the adversarial phase.

\textbf{Policy initialization.} The policy is initialized from the SFT checkpoint described in Section~\ref{sec:sft}, which narrows the gap between policy rollouts and the supervision distribution sufficiently for adversarial training to begin.

\textbf{MoE discriminator initialization.} Both experts are initialized from the same pretrained backbone and warm-started on their designated components: $D_v$ on preference pairs from visual descriptions, $D_r$ on preference pairs from reasoning traces. An auxiliary load-balancing loss~\citep{fedus2022switch} prevents expert collapse during this stage.

\subsubsection{Adversarial On-Policy Distillation}

With all components properly initialized, we formulate the alignment stage as a minimax game between the policy $G$ and the MoE discriminator. The policy is optimized to generate responses that increasingly resemble high-quality reference demonstrations, while the discriminator is trained to separate the two. Through this adversarial interaction, the policy distribution is progressively driven toward the reference distribution, yielding a more faithfully aligned model before RLVR.

Specifically, the MoE discriminator assigns a scalar score to each response based on both perceptual grounding and reasoning consistency. Let $\mathcal{T}$ denote the supervision data from which reference pairs are drawn. Given a policy response $y^{-}$ and a reference response $y^{+}$, we train the discriminator to assign a higher score to the reference and a lower score to the policy rollout by minimizing its Bradley-Terry loss:
\begin{compactmath}
\begin{equation}
\mathcal{L}_{D_k}=-\mathbb{E}_{(x,y^{+},y^{-}) \sim \mathcal{T}}\left[\log\sigma\bigl(D_k(x, y^{+}_{k}) - D_k(x, y^{-}_{k}) \bigr)\right],\qquad k \in \{v,r\},
\end{equation}
\end{compactmath}
where $y^{+}_{k}$ and $y^{-}_{k}$ denote the $k$-th component of the reference response $y^{+}$ and the policy response $y^{-}$, respectively, with $k=v$ corresponding to the visual description and $k=r$ corresponding to the reasoning trace. Here, $y^{-} \sim G(\cdot \mid x)$ is sampled from the current policy, and $\sigma(\cdot)$ is the sigmoid function. Importantly, both experts are optimized jointly with the policy throughout alignment, so that they function as on-policy discriminators that continuously adapt to the evolving rollout distribution. This design avoids the reward staleness issue that commonly arises when the reward model is fixed while the policy keeps changing.

The policy is optimized to improve the quality of its own rollouts under the reward provided by the MoE discriminator. For each input $x$, we sample a group of $N$ responses $\{y^{-}_{i}\}_{i=1}^{N}$ from the current policy $G(\cdot \mid x)$, and evaluate each response with the discriminator reward $r(x, y^{-}_{i})$. We convert these rewards into normalized group-wise advantages:
\begin{compactmath}
\begin{equation}
A_i=\frac{r(x, y^{-}_{i}) - \operatorname{mean}\!\left(\{r(x,y^{-}_{j})\}_{j=1}^{N}\right)}{\operatorname{std}\!\left(\{r(x,y^{-}_{j})\}_{j=1}^{N}\right)},
\end{equation}
\end{compactmath}
where the normalization is performed within each rollout group. In this way, the policy is encouraged to increase the probability of responses that are scored as more consistent with the supervision distribution, while suppressing inferior rollouts from the same prompt. Taken together, the two objectives define a minimax game between the policy and the discriminator:
\begin{compactmath}
\begin{equation}
\min_{\theta}\max_{\phi}\;
\mathbb{E}_{(x,y^{+}) \sim \mathcal{T},\, y^{-}\sim G_{\theta}(\cdot \mid x)}
\Big[r_{\phi}(x,y^{+}) - r_{\phi}(x,y^{-})\Big],
\end{equation}
\end{compactmath}
where $\theta$ and $\phi$ denote the parameters of the policy and discriminator, respectively.

We alternate between updating the policy via GRPO and updating the two discriminator experts with their respective Bradley-Terry losses. Notably, we remove the KL regularization term commonly used to anchor the policy near its SFT initialization, since such a constraint would directly oppose the goal of correcting SFT-induced distributional drift. The alignment stage is run for a fixed number of steps, after which the resulting checkpoint is used to initialize the final RLVR stage.

\subsection{Reinforcement Learning with Verifiable Rewards}
\label{sec:rlvr}
The final stage of PRISM applies standard outcome-based RLVR to the aligned checkpoint produced in Section~\ref{sec:alignment}. To construct the RL training set, we filter the 6K reserved samples by difficulty, retaining the 2K instances whose pass rate under the aligned policy falls in $[0.2, 0.8]$~\citep{wang2025reinforcement,zhang2026resource}. At this stage, the reward switches from the learned MoE discriminator to a deterministic verifiable reward. The reward combines answer accuracy $r_{\text{acc}}$ and format compliance $r_{\text{fmt}}$:
\begin{compactmath}
\begin{equation}
r_{\text{v}}(x,y) = 0.8\, r_{\text{acc}}(x,y) + 0.2\, r_{\text{fmt}}(x,y).
\end{equation}
\end{compactmath}
Policy optimization follows the same GRPO-style objective as in the alignment stage, with the discriminator reward replaced by $r_{\text{v}}$. Importantly, PRISM is agnostic to the specific RL algorithm; in our experiments, we instantiate this stage with GRPO~\citep{shao2024deepseekmath}, DAPO~\citep{yu2025dapo}, and GSPO~\citep{zheng2025group}.

\section{Experiments}
\label{sec:experiments}

\subsection{Experimental Setup}
\label{sec:setup}

\textbf{Models.}
We use Qwen3-VL-4B and Qwen3-VL-8B~\citep{bai2025qwen3} as the base models, and Gemini~3 Flash as the supervision source for data construction. The MoE discriminator follows the Qwen3-VL-MoE architecture~\citep{bai2025qwen3}, constructed by assembling four Qwen3-VL-2B models as expert modules with top-2 routing. The two expert scores $D_v$ and $D_r$ are obtained by passing the visual description and reasoning trace through this single MoE model separately, then combined via Eq.~\eqref{eq:moe_reward}.

\textbf{Training details.} We follow the three-stage pipeline described in Section~\ref{sec:method}: SFT for 1 epoch on the full supervision dataset, distribution alignment for 500 steps, and RLVR for 1500 steps with verifiable rewards. Detailed hyperparameters for all stages (learning rates, batch sizes, $\alpha$, rollout temperature, group size $N$, etc.) are provided in Appendix~\ref{app:hyperparams}.

\textbf{Benchmarks.}
We evaluate on two groups of benchmarks. For mathematical reasoning, we use MathVista~\citep{lu2023mathvista}, MathVerse~\citep{zhang2024mathverse}, MathVision~\citep{wang2024measuring}, and WeMath~\citep{qiao2025we}. For general multimodal understanding, we use MMMU~\citep{yue2024mmmu}, MMMU-Pro~\citep{yue2025mmmu}, and HallusionBench~\citep{guan2024hallusionbench}.

\textbf{Baselines.}
We compare against the following: (1) Qwen3-VL models without post-training, (2) SFT-only models trained on the full dataset, and (3) the standard SFT$\rightarrow$RLVR pipeline without the alignment stage, using GRPO~\citep{shao2024deepseekmath} and trained with additional RLVR steps equal to the alignment stage to match the number of policy optimization updates. For the PRISM pipeline, we report results with GRPO, DAPO~\citep{yu2025dapo}, and GSPO~\citep{zheng2025group} in the RLVR stage to demonstrate algorithm-agnostic improvements.

\subsection{Main Results}

\begin{table}[t]
\caption{Accuracy (\%) on mathematical reasoning and general multimodal benchmarks. \textbf{Bold} indicates the best result within each base model. \colorbox{gray!10}{Shaded rows} denote PRISM (ours).}
\label{tab:main}
\vspace{-4pt}
\centering
\small
\resizebox{\linewidth}{!}{
\begin{tabular}{l cccc ccc c}
\toprule
& \multicolumn{4}{c}{\textbf{Math Benchmarks}} & \multicolumn{3}{c}{\textbf{General Benchmarks}} & \\
\cmidrule(lr){2-5} \cmidrule(lr){6-8}
\textbf{Method} & \makecell{\textbf{MathVista}\\testmini} & \makecell{\textbf{MathVerse}\\testmini} & \makecell{\textbf{MathVision}\\testmini} & \makecell{\textbf{WeMath}\\testmini} & \makecell{\textbf{MMMU}\\val} & \makecell{\textbf{MMMU-Pro}\\overall} & \makecell{\textbf{Hallusion}\\\textbf{Bench}} & \textbf{Avg.} \\
\midrule
\multicolumn{9}{c}{\textit{\textbf{Qwen3-VL-4B}}} \\
\midrule
Instruct & 74.9 & 59.0 & 36.5 & 70.7 & 63.6 & 45.1 & 68.2 & 59.7 \\
\quad + SFT & 71.5 & 58.4 & 31.9 & 70.6 & 53.6 & 42.8 & 69.1 & 56.8 \\
\quad + GRPO & 75.7 & 64.5 & 35.5 & 77.8 & 60.1 & 47.3 & 72.0 & 61.8 \\
\quad + DAPO & 74.3 & 65.1 & 42.7 & 77.2 & 62.5 & 48.0 & 72.3 & 63.2 \\
\quad + GSPO & 75.2 & 64.0 & 37.5 & 78.4 & 58.7 & 45.6 & 71.9 & 61.6 \\
\cdashline{1-9}
\rowcolor{gray!10} PRISM & 71.0 & 59.5 & 30.6 & 67.5 & 56.3 & 42.8 & 72.6 & 57.2 \\
\rowcolor{gray!10} \quad + GRPO & \textbf{77.9} & \textbf{68.6} & 45.4 & 82.9 & \textbf{64.1} & 49.7 & \textbf{74.8} & 66.2 \\
\rowcolor{gray!10} \quad + DAPO & 77.8 & 68.2 & \textbf{46.7} & \textbf{83.9} & \textbf{64.1} & 50.4 & 72.9 & \textbf{66.3} \\
\rowcolor{gray!10} \quad + GSPO & 77.5 & 66.6 & \textbf{46.7} & 82.3 & 63.2 & \textbf{51.1} & 72.9 & 65.8 \\
\midrule
\multicolumn{9}{c}{\textit{\textbf{Qwen3-VL-8B}}} \\
\midrule
Instruct & 76.0 & 62.4 & 43.7 & 71.7 & 65.6 & 52.3 & 71.6 & 63.3 \\
\quad + SFT & 70.2 & 60.4 & 32.6 & 73.4 & 56.3 & 42.9 & 71.2 & 58.1 \\
\quad + GRPO & 75.9 & 66.9 & 37.1 & 79.7 & 62.6 & 48.8 & 71.9 & 63.3 \\
\quad + DAPO & 77.0 & 69.8 & 41.5 & 84.3 & 63.0 & 49.0 & 71.5 & 65.2 \\
\quad + GSPO & 75.9 & 65.5 & 41.1 & 80.8 & 58.2 & 47.8 & 73.6 & 63.3 \\
\cdashline{1-9}
\rowcolor{gray!10} PRISM & 71.4 & 62.2 & 37.1 & 73.1 & 58.4 & 43.4 & 69.5 & 59.3 \\
\rowcolor{gray!10} \quad + GRPO & \textbf{78.3} & 71.3 & \textbf{52.0} & \textbf{86.4} & \textbf{66.6} & \textbf{53.3} & \textbf{77.2} & \textbf{69.3} \\
\rowcolor{gray!10} \quad + DAPO & 78.2 & 70.9 & \textbf{52.0} & 86.2 & 66.2 & 52.4 & 76.1 & 68.9 \\
\rowcolor{gray!10} \quad + GSPO & 77.9 & \textbf{71.5} & 51.6 & 85.9 & 65.2 & 52.7 & 75.8 & 68.7 \\
\bottomrule
\end{tabular}
}
 \vspace{-4mm}
\end{table}

Table~\ref{tab:main} presents results across both model scales and three RL algorithms. We highlight three key observations.

\textbf{PRISM consistently improves downstream RLVR.} PRISM+GRPO outperforms the SFT$\rightarrow$GRPO baseline by +4.4 and +6.0 average points on 4B and 8B, respectively, with the largest gains concentrated on MathVision and WeMath across both scales. Similar improvements hold for DAPO and GSPO, confirming that the alignment stage provides a consistently better initialization regardless of the downstream RL algorithm. Moreover, PRISM+GRPO achieves these gains while using fewer tokens per response (Appendix~\ref{app:token}).

\textbf{Alignment improves distribution, not immediate accuracy.} The PRISM row, representing the checkpoint after alignment but before RLVR, shows accuracy comparable to the SFT checkpoint on both scales. This is expected: the alignment objective is distributional correction, reshaping the model's visual descriptions and reasoning traces to better match the supervision distribution, rather than directly optimizing for answer correctness. The value of this distributional shift is realized downstream, where every RL algorithm benefits from the improved initialization.

\textbf{SFT-induced drift is more pronounced for stronger models.} Consistent with the observation discussed in the introduction, SFT degrades the Instruct checkpoint on average across both scales, with the 8B model suffering a larger drop than 4B. On 8B, the standard SFT$\rightarrow$RLVR pipeline with GRPO and GSPO barely recovers the original Instruct performance, indicating that RLVR alone cannot fully compensate for the distributional drift introduced by SFT on a model with an already-strong prior. In contrast, PRISM+GRPO exceeds the Instruct baseline by over 5 points, demonstrating that the alignment stage effectively repairs the drift and unlocks further gains from RLVR.

\subsection{Ablation Study}
\label{sec:ablation}
All ablations are conducted on Qwen3-VL-4B with GRPO as the default RL algorithm.

\begin{table}[htbp]
\caption{Ablation study results. The first row is the full PRISM pipeline (SFT-1.37M) for reference.}
\label{tab:ablation}
\centering
\small
\resizebox{\linewidth}{!}{
\begin{tabular}{l cccc ccc c}
\toprule
& \multicolumn{4}{c}{\textbf{Math Benchmarks}} & \multicolumn{3}{c}{\textbf{General Benchmarks}} & \\
\cmidrule(lr){2-5} \cmidrule(lr){6-8}
\textbf{Setting} & \makecell{\textbf{MathVista}\\testmini} & \makecell{\textbf{MathVerse}\\testmini} & \makecell{\textbf{MathVision}\\testmini} & \makecell{\textbf{WeMath}\\testmini} & \makecell{\textbf{MMMU}\\val} & \makecell{\textbf{MMMU-Pro}\\overall} & \makecell{\textbf{Hallusion}\\\textbf{Bench}} & \textbf{Avg.} \\
\midrule
PRISM (full) & 77.9 & 68.6 & 45.4 & 82.9 & 64.1 & 49.7 & 74.8 & 66.2 \\
\midrule
\multicolumn{9}{c}{\textit{Discriminator design}} \\
\midrule
Dense 4B disc. & 74.6 & 63.7 & 41.8 & 76.9 & 61.3 & 47.1 & 74.0 & 62.8 \\
Text-only disc. & 74.0 & 59.5 & 42.8 & 76.8 & 62.7 & 48.5 & 71.6 & 62.3 \\
\midrule
\multicolumn{9}{c}{\textit{Pipeline stages}} \\
\midrule
w/o SFT & 62.4 & 47.6 & 25.9 & 55.7 & 51.4 & 36.5 & 66.1 & 49.4 \\
w/o Alignment & 75.7 & 64.5 & 35.5 & 77.8 & 60.1 & 47.3 & 72.0 & 61.8 \\
\midrule
\multicolumn{9}{c}{\textit{SFT data scale}} \\
\midrule
SFT-107K & 72.3 & 67.0 & 43.1 & 76.9 & 60.6 & 49.0 & 68.3 & 62.5 \\
\bottomrule
\end{tabular}
}
 \vspace{-1mm}
\end{table}

\textbf{Why does the MoE discriminator matter?}
As shown in Table~\ref{tab:ablation}, replacing the MoE discriminator with a single dense model with equal activated compute leads to consistent degradation ($-3.4$ avg.), with the largest drops on WeMath ($-6.0$) and MathVerse ($-4.9$). The dense discriminator collapses perception and reasoning feedback into a single scalar: when the policy improves along one axis but regresses along the other, the monolithic reward cannot disentangle the two effects and the gradient signal becomes noisy. The MoE design avoids this by letting each expert specialize on its own domain. The training dynamics in Section~\ref{sec:analysis} further confirm that the two experts converge along distinct trajectories, a structure that a single discriminator would inevitably conflate.

\textbf{Why is the three-stage pipeline necessary?}
We ablate each training stage individually and find that strong performance emerges only with the full three-stage pipeline, consistent with the observation in~\citep{yang2025longvt} that each stage in a multi-stage post-training pipeline serves an indispensable role. Removing the alignment stage reduces the pipeline to standard SFT$\rightarrow$RLVR with a $-4.4$ avg. drop, indicating that without explicit distribution correction, RLVR alone cannot fully compensate for the distributional drift inherited from SFT. Removing SFT leads to an even starker picture ($-16.8$ avg.): without cold-start initialization, the capability gap between the model and the supervision distribution is too large for adversarial training to function: the discriminator trivially separates the two distributions, and the policy drifts toward a degenerate output mode rather than converging toward the supervision distribution. Each stage thus serves an indispensable role: SFT provides initial competence and narrows the distribution gap enough for adversarial training to begin; alignment refines what the reasoning should look like by explicitly closing the residual gap with the supervision distribution; and RLVR optimizes whether the reasoning leads to a correct answer.

\textbf{Why must the discriminator be vision-language?}
A natural question is whether the discriminator truly needs to see the image, or whether textual patterns alone carry enough signal. We replace the vision-language discriminator with a text-only variant of identical architecture. The text-only variant can capture surface-level textual patterns such as formatting conventions, reasoning templates, and stylistic cues, but it cannot verify whether a visual description faithfully depicts the image content. The result is a form of ``parrot alignment'': the policy learns to sound like the supervision data without actually seeing what it describes. Consistently, the degradation is most pronounced on tasks that demand faithful visual perception, confirming that meaningful distribution alignment in the multimodal setting requires a discriminator that jointly processes vision and language.

\textbf{How important is SFT data scale?}
When using only our curated 107K samples for SFT instead of the full 1.37M corpus (107K curated + 1.26M public), downstream RLVR performance drops by $-3.7$ avg. With less SFT supervision, the model starts further from the target distribution, widening the gap that the alignment stage must bridge. While alignment can partially compensate, as the reduced-data variant (62.5 avg.) still outperforms the SFT$\rightarrow$RLVR baseline trained on the full corpus but without alignment (61.8 avg.), it cannot fully recover from a weaker initialization. This confirms that SFT and alignment are complementary rather than substitutable: broad SFT data tightens the initial distribution gap, and alignment closes whatever residual gap remains.

\subsection{Analysis}
\label{sec:analysis}
\begin{wrapfigure}{r}{0.5\textwidth}
    \centering
    \vspace{-10pt}
    \includegraphics[width=0.5\textwidth]{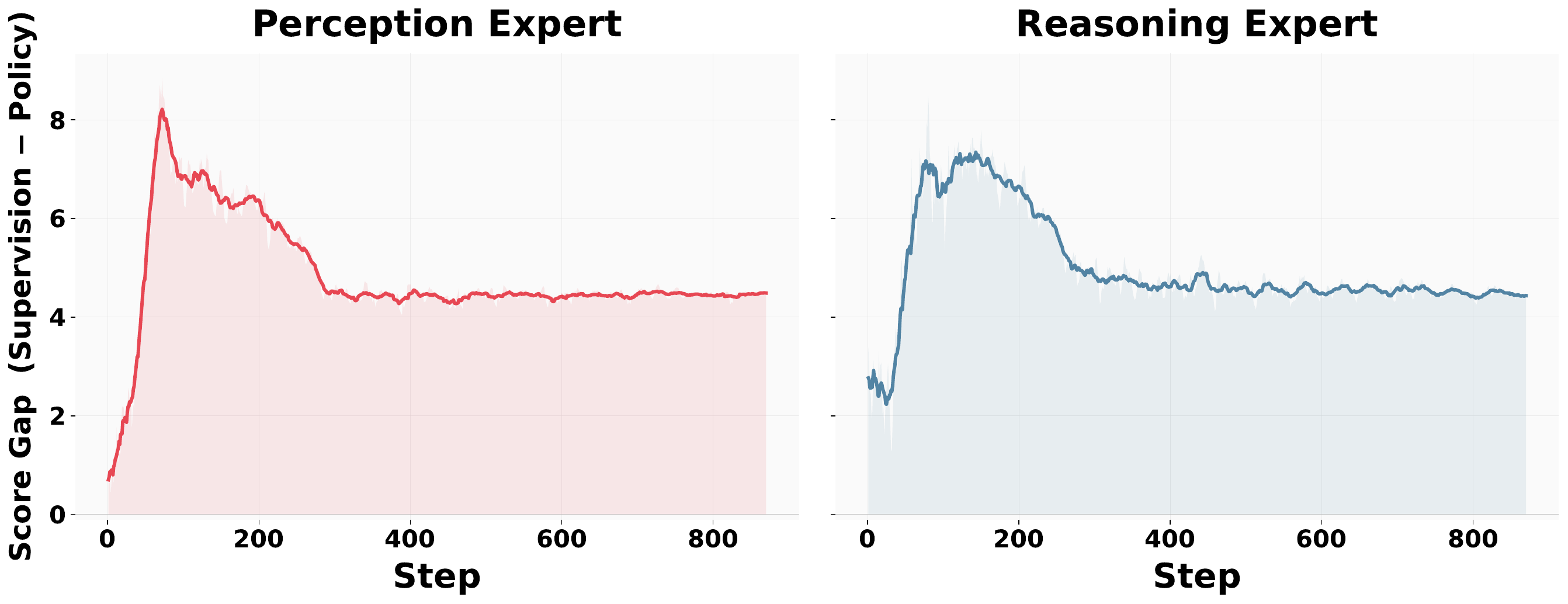}
    \vspace{-8pt}
    \caption{Training dynamics: reward gap (supervision $-$ policy) for the perception expert (left) and reasoning expert (right). Training runs for 500 steps; we additionally extend to 900 steps to verify convergence stability.}
    \label{fig:dynamics}
    \vspace{-10pt}
\end{wrapfigure}
\textbf{Training dynamics.}
We analyze the adversarial training dynamics of PRISM during the alignment stage by tracking the reward gap between supervision and policy responses, defined as $D_k(x, y^{+}_k) - D_k(x, y^{-}_k)$ for each expert $k \in \{v, r\}$. As shown in Figure~\ref{fig:dynamics}, both experts exhibit a characteristic rise-then-convergence pattern, but with notably different trajectories that reflect the distinct nature of visual and reasoning alignment.

The perception expert peaks early and converges quickly, whereas the reasoning expert rises more gradually and exhibits greater oscillation before stabilizing, suggesting that reasoning alignment involves more nuanced distributional corrections than visual grounding. Despite these different trajectories, both experts converge to a comparable equilibrium level, confirming that the adversarial game reaches a stable state where the policy has been substantially realigned toward the supervision distribution along both axes.

This asynchronous convergence further supports the MoE design choice validated in the ablation study (Section~\ref{sec:ablation}).

\vspace{-0.1mm}
\textbf{Structural Proxies of Distribution Alignment.}
Since directly visualizing the full multimodal response distribution is difficult in such high-dimensional output spaces, we instead probe its evolution through interpretable structural proxies~\citep{tang2016visualizing}. This choice is also consistent with prior work that analyzes model-generated reasoning traces through their structural qualities and patterns~\citep{mathur2025social,jiang2025makes}. Concretely, we focus on the two dimensions most relevant to PRISM: reasoning complexity and perceptual grounding granularity. We visualize the number of reasoning steps in the chain-of-thought trace and the number of descriptive items in the visual caption. We compute these statistics for the base model, the supervision data, and the policy checkpoints after SFT, alignment, and RLVR.

\begin{wrapfigure}{r}{0.5\textwidth}
    \centering
    \vspace{-10pt}
    \includegraphics[width=0.5\textwidth]{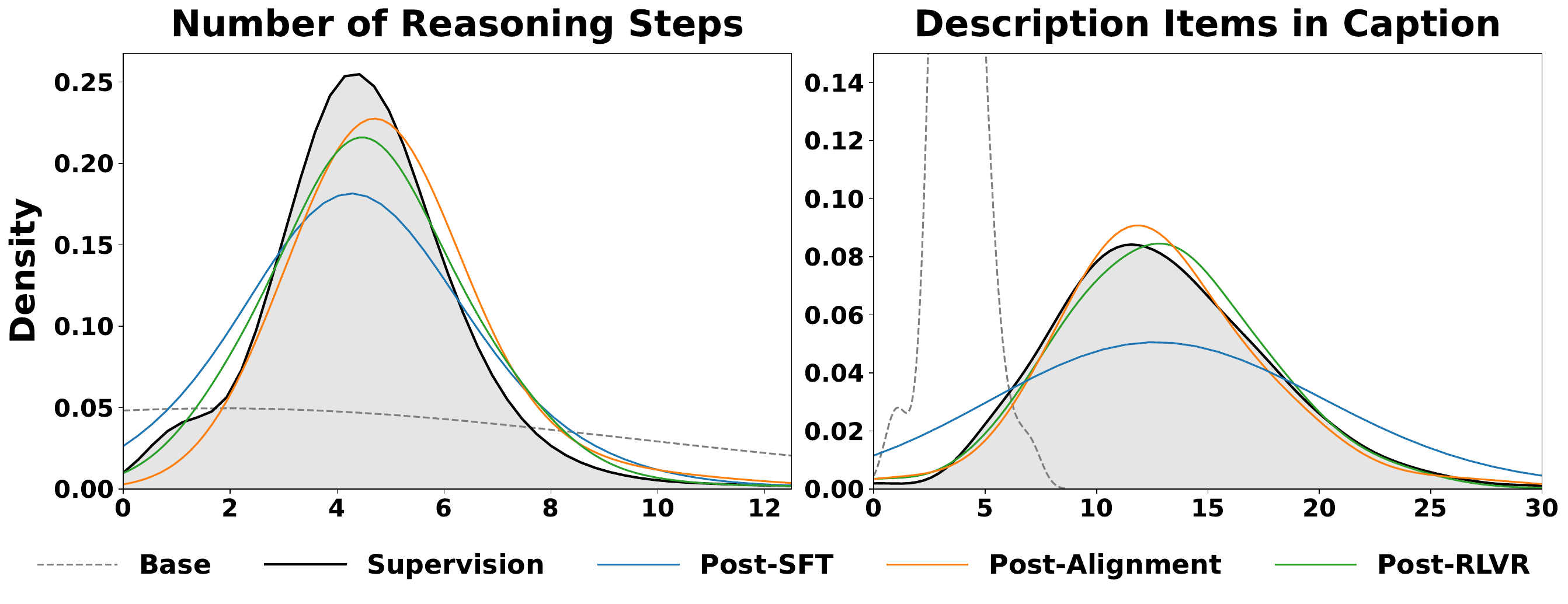}
    \vspace{-10pt}
    \caption{Structural proxies of distribution alignment: reasoning steps (left) and descriptive items per caption (right) across PRISM stages.}
    \label{fig:distribution}
    \vspace{-8pt}
\end{wrapfigure}
As shown in Figure~\ref{fig:distribution}, the base model deviates substantially from the supervision data under both proxies. After SFT, the response distributions shift toward supervision but remain visibly mismatched, particularly because the caption-side proxy tends to overshoot and produce more descriptive items than the supervision data contains. The alignment stage substantially reduces this mismatch along both dimensions. Importantly, this improvement persists through RLVR: the post-RLVR distributions continue to follow similar trends despite the shift in optimization objective from distributional matching to outcome-based correctness. This suggests that the alignment stage provides a better-shaped initialization for subsequent RLVR, which further improves performance while preserving the alignment gains.

\section{Conclusion}
\label{sec:conclusion}

We present PRISM, a three-stage post-training pipeline that mitigates the distributional drift introduced by SFT through an explicit alignment stage based on black-box adversarial on-policy distillation. The core of PRISM is an MoE discriminator with dedicated perception and reasoning experts, which provides disentangled corrective signals to steer the post-SFT policy toward the supervision distribution under the model's own rollout dynamics. Extensive experiments on Qwen3-VL demonstrate that PRISM consistently improves downstream RLVR performance across multiple RL algorithms and diverse multimodal benchmarks. Analysis of training dynamics and response-structure distributions suggests that the alignment stage substantially narrows the distributional gap left by SFT, and that these corrections persist through subsequent RL optimization.

\section*{Acknowledgments}
This work is supported by funding from ModelBest.

\bibliographystyle{unsrtnat}
\bibliography{references}

\newpage
\appendix
%\section*{Appendix}

\section{Implementation Details}
\label{app:hyperparams}

\subsection{Training Details}

We describe the training procedures for all three stages of the PRISM pipeline. The SFT stage is implemented using LlamaFactory~\citep{zheng2024llamafactory}, while the alignment and RLVR stages are both built on the veRL framework~\citep{sheng2025hybridflow} with vLLM~\citep{kwon2023efficient} as the inference engine. The full set of hyperparameters is provided in Table~\ref{tab:hyperparams}.

\begin{table}[h]
\caption{Detailed training hyperparameters for each stage of the PRISM pipeline.}
\label{tab:hyperparams}
\centering
\small
\begin{tabular}{l ccc}
\toprule
\textbf{Training Component} & \textbf{SFT} & \textbf{PRISM} & \textbf{RLVR} \\
\midrule
Optimizer & AdamW & AdamW & AdamW \\
Scheduler & cosine & constant & constant \\
Learning Rate & 1e-5 & 1e-6 & 1e-6 \\
Weight Decay & -- & 0.01 & 0.01 \\
Epochs / Training Steps & 1 epoch & 500 steps & 1500 steps \\
Warmup Ratio / Steps & 0.1 & -- & -- \\
Global Batch Size & 2 & 4 & 32 \\
Max Prompt Length & -- & 2048 & 2048 \\
Max Response Length & 8192 & 6144 & 8192 \\
Rollout Temperature & -- & 1.0 & 1.0 \\
Group Size $N$ & -- & 16 & 16 \\
$\alpha$ (MoE weight) & -- & 0.5 & -- \\
Accuracy Reward Weight & -- & -- & 0.8 \\
Format Reward Weight & -- & -- & 0.2 \\
Dynamic Bsz & -- & True & True \\
Remove Padding & -- & True & True \\
\bottomrule
\end{tabular}
\end{table}

\paragraph{SFT.}
We perform full-parameter fine-tuning on the language model while freezing the vision tower and the multimodal projector. We use a cutoff length of 8192 tokens with online stream packing to maximize training throughput and remove padding tokens to avoid unnecessary computation. Training is conducted for 1 epoch with a cosine learning rate schedule (peak 1e-5, warmup ratio 0.1) using DeepSpeed ZeRO-2.

\paragraph{PRISM (Alignment).}
The alignment stage jointly trains the policy and the MoE discriminator. The generator and discriminator share the same learning rate (1e-6). For each prompt, we sample $N=16$ rollouts from the current policy with temperature 1.0. The MoE reward weight is set to $\alpha=0.5$, giving equal importance to the perception and reasoning experts. We deliberately disable KL regularization (KL coefficient = 0.0) to allow the policy to shift freely toward the supervision distribution. 
% We train for 500 steps, at which point both discriminator experts have converged to a stable equilibrium (see Figure~\ref{fig:dynamics}), and select the final checkpoint for the subsequent RLVR stage.
We train for 500 steps, at which point both discriminator experts have converged to a stable equilibrium (see Figure~\ref{fig:dynamics}), and select the final checkpoint for the subsequent RLVR stage.
% caption_weight=0.5, cot_weight=0.5, format_penalty=1.5, score_clip=10.0, score_reg_coef=0.001

\paragraph{RLVR.}
Starting from the alignment checkpoint, we apply outcome-based RLVR with a global batch size of 32 and $N=16$ rollouts per prompt at temperature 1.0. All RL algorithms evaluated in our experiments share the same set of hyperparameters. The maximum response length is increased to 8192 tokens to accommodate longer reasoning traces. We train for up to 1500 steps, saving checkpoints periodically and selecting the best one based on validation performance. All experiments across all three stages are conducted on 8$\times$ NVIDIA H100-SXM5-80GB GPUs.

\subsection{Evaluation Details}
\label{app:eval}

For all evaluations, we use the lmms-eval framework~\citep{zhang2025lmms} with vLLM~\citep{kwon2023efficient} as the serving engine. We set the generation hyperparameters to \texttt{temperature=1.0}, \texttt{top\_p=0.7}, and \texttt{top\_k=-1} across all benchmarks. For MathVerse and HallusionBench, we follow their official protocols, which use an LLM judge for every response. For the remaining benchmarks, the extracted answer is validated using a two-stage process: a rule-based validator is first applied to minimize evaluation cost; if the rule-based validator marks the answer as incorrect, we fall back to an LLM-as-judge validator powered by Qwen3-30B-A3B-Instruct~\citep{yang2025qwen3} with a conservative binary judging prompt (Figure~\ref{fig:prompt-judge}). All methods share the same evaluation pipeline.

\paragraph{Judge usage and robustness.} Table~\ref{tab:judge_usage} reports the fraction of responses handled by the LLM judge on each benchmark. Pooled over all evaluated responses, judge usage is 37.1\% on the rule-first settings and 58.6\% overall. To assess sensitivity to the choice of judge, we re-judge all judge-handled responses with Qwen3-235B-A22B-Instruct~\citep{yang2025qwen3}; aggregate scores change by less than 0.1 percentage points, and the ranking of methods and their relative gains are unchanged.

\begin{table}[!htbp]
\caption{Fraction of responses handled by the LLM judge on each evaluation setting.}
\label{tab:judge_usage}
\centering
\small
\begin{tabular}{lc}
\toprule
\textbf{Evaluation setting} & \textbf{Judge usage} \\
\midrule
MathVista testmini & 27.4\% \\
MathVision testmini & 58.2\% \\
MMMU val & 39.6\% \\
MMMU-Pro standard & 49.4\% \\
MMMU-Pro vision & 53.9\% \\
WeMath testmini & 20.0\% \\
MathVerse testmini (official LLM-judge protocol) & 100\% \\
HallusionBench (official LLM-judge protocol) & 100\% \\
\bottomrule
\end{tabular}
\end{table}

\subsection{Token Efficiency}
\label{app:token}

\begin{figure}[h]
    \centering
    \includegraphics[width=\textwidth]{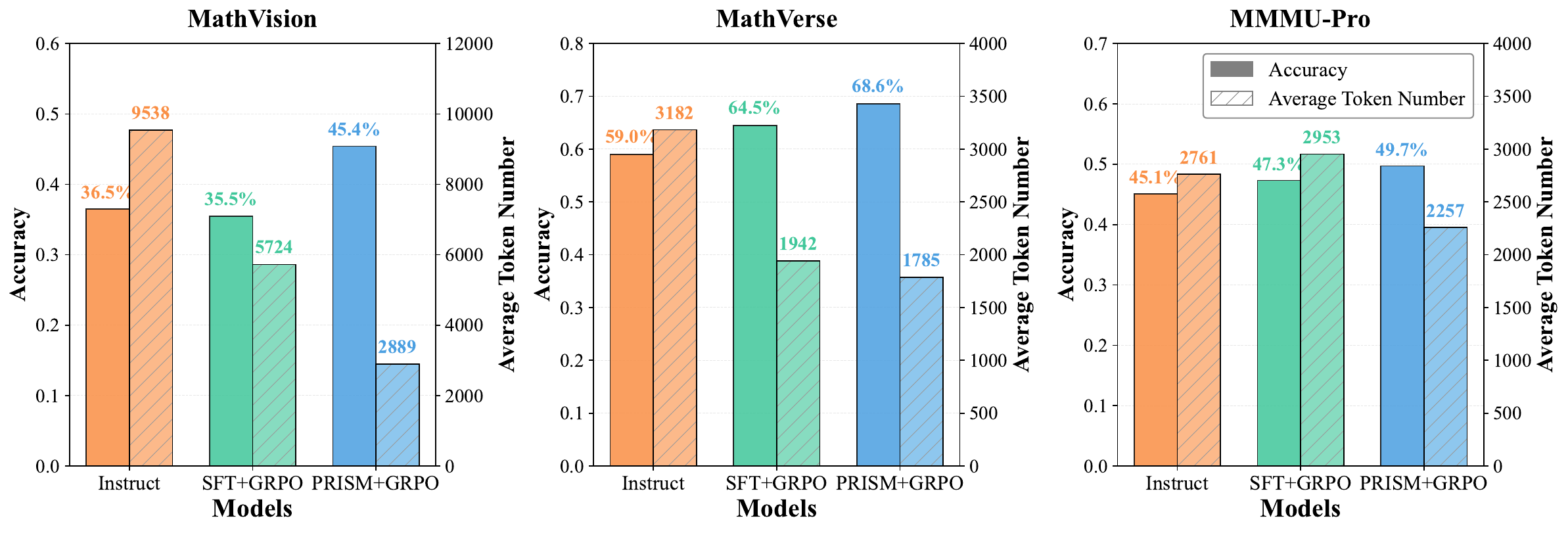}
    \caption{Token efficiency comparison on MathVision, MathVerse, and MMMU-Pro (Qwen3-VL-4B). PRISM+GRPO achieves higher accuracy with fewer tokens across all three benchmarks.}
    \label{fig:token_efficiency}
\end{figure}

Figure~\ref{fig:token_efficiency} compares the accuracy and average token usage of three configurations on MathVision, MathVerse, and MMMU-Pro. Across all three benchmarks, PRISM+GRPO achieves higher accuracy than SFT+GRPO while using fewer tokens, suggesting that the alignment stage encourages more concise and effective reasoning rather than simply producing longer outputs.

\section{Additional Controlled Experiments}
\label{app:controls}

This section examines whether the gains of PRISM can be attributed to the pre-alignment stage rather than to longer training, additional data, or the choice of objective. Owing to compute constraints, all experiments in this section use Qwen3-VL-4B with GRPO and are single runs. All RLVR runs use the same 2K RLVR subset and rollout configuration; except for Instruct and Instruct$\rightarrow$GRPO, all methods start from the same SFT checkpoint. Results are summarized in Table~\ref{tab:controls}.

\begin{table}[!htbp]
\caption{Controlled comparisons on Qwen3-VL-4B (GRPO). Where shown, update counts are given as alignment / RLVR policy updates. \textbf{Bold} indicates the best result in each column.}
\label{tab:controls}
\centering
\small
\resizebox{\linewidth}{!}{
\begin{tabular}{l cccc ccc c}
\toprule
& \multicolumn{4}{c}{\textbf{Math Benchmarks}} & \multicolumn{3}{c}{\textbf{General Benchmarks}} & \\
\cmidrule(lr){2-5} \cmidrule(lr){6-8}
\textbf{Method} & \textbf{MathVista} & \textbf{MathVerse} & \textbf{MathVision} & \textbf{WeMath} & \textbf{MMMU} & \textbf{MMMU-Pro} & \textbf{HallusionBench} & \textbf{Avg.} \\
\midrule
Instruct (no post-training) & 74.9 & 59.0 & 36.5 & 70.7 & 63.6 & 45.1 & 68.2 & 59.7 \\
\midrule
\multicolumn{9}{c}{\textit{Longer RLVR training of the baseline}} \\
\midrule
SFT$\rightarrow$GRPO (0 / 2,000) & 75.7 & 64.5 & 35.5 & 77.8 & 60.1 & 47.3 & 72.0 & 61.8 \\
SFT$\rightarrow$GRPO (0 / 3,000) & 76.4 & 64.7 & 35.8 & 78.1 & 59.6 & 46.3 & 71.4 & 61.8 \\
SFT$\rightarrow$GRPO (0 / 5,000) & 74.6 & 64.3 & 33.1 & 82.3 & 57.8 & 45.7 & 71.0 & 61.3 \\
\midrule
\multicolumn{9}{c}{\textit{Alternative initialization and objective}} \\
\midrule
Instruct$\rightarrow$GRPO & 77.5 & 65.6 & 39.8 & 77.1 & 63.9 & 48.6 & 73.9 & 63.8 \\
VOLD-style joint OPD+RL & 75.8 & 63.2 & 34.0 & 76.7 & 60.7 & 46.9 & 72.4 & 61.4 \\
\midrule
\multicolumn{9}{c}{\textit{PRISM (staged)}} \\
\midrule
PRISM-2K$\rightarrow$GRPO (500 / 1,500) & 77.7 & 68.3 & 44.9 & 82.6 & \textbf{64.3} & \textbf{49.8} & 74.6 & 66.0 \\
PRISM$\rightarrow$GRPO (500 / 1,500) & \textbf{77.9} & \textbf{68.6} & \textbf{45.4} & \textbf{82.9} & 64.1 & 49.7 & \textbf{74.8} & \textbf{66.2} \\
\bottomrule
\end{tabular}
}
\end{table}

\paragraph{Matched updates and longer baseline training.} SFT$\rightarrow$GRPO uses 2,000 RLVR updates, while PRISM uses 500 alignment and 1,500 RLVR updates, so the two pipelines are matched in the number of policy optimization updates but not in FLOPs, since PRISM additionally trains the MoE discriminator. To test whether the baseline is simply under-trained, we extend SFT$\rightarrow$GRPO to 3,000 and 5,000 updates with the same data and settings. Its average accuracy plateaus at 61.8 and drops to 61.3 at 5,000 updates: WeMath improves, but most other benchmarks degrade, suggesting over-optimization on the RLVR training set. PRISM remains 4.4 points above the best extended baseline while using 2,000 total policy updates, so within the tested horizon the gap is not explained by insufficient baseline training.

\paragraph{Data-matched alignment.} PRISM samples alignment prompts from the 6K held-out pool, but its 500 alignment updates with batch size 4 amount to only 2K prompt exposures, so the 6K set mainly provides a larger sampling pool rather than more training exposure. We further train PRISM-2K, which uses the identical 2K subset for both alignment and RLVR, so that the set of unique prompts matches the baseline. PRISM-2K reaches 66.0, only 0.2 points below the 6K-pool variant and 4.2 points above SFT$\rightarrow$GRPO, indicating that the gain is not primarily due to the larger alignment pool.

\paragraph{Direct RLVR on the Instruct model.} Applying GRPO directly to Qwen3-VL-4B-Instruct on the same 2K subset with the same configuration outperforms SFT$\rightarrow$GRPO by 2.0 points, consistent with the observation that SFT can partially suppress useful capabilities of the Instruct model. However, Instruct$\rightarrow$GRPO remains 2.4 points below PRISM$\rightarrow$GRPO and is lower on all seven benchmarks, so PRISM does more than recover the original Instruct capability. Note that applying alignment directly to the Instruct model without SFT (``w/o SFT'' in Table~\ref{tab:ablation}) performs much worse, consistent with our hypothesis in Section~\ref{sec:ablation} that adversarial alignment requires an initialization that is already reasonably close to the supervision distribution.

\paragraph{Comparison with joint OPD+RL.} The closest related method, VOLD~\citep{bousselham2025vold}, combines on-policy distillation and RL in a single objective, but requires white-box teacher token probabilities that are unavailable from Gemini. We therefore implement a black-box joint variant that jointly optimizes the discriminator reward and the GRPO correctness reward, using the same initialization, data, discriminator, rollout configuration, and policy-update budget as staged PRISM. This joint variant (61.4) underperforms both SFT$\rightarrow$GRPO and staged PRISM, suggesting interference between the distribution-matching and correctness objectives: GRPO rewards answer correctness, whereas the discriminator rewards resemblance to the supervision distribution, so optimizing both simultaneously may produce conflicting updates. PRISM avoids this conflict by restricting discriminator feedback to the pre-alignment stage and using only verifiable rewards during RLVR. Other related post-training approaches differ in key respects: offline preference-optimization objectives such as DPO~\citep{rafailov2023direct} operate on fixed response pairs, while RLAIF~\citep{lee2023rlaif} optimizes the policy against a reward model trained on AI-labeled preferences rather than against a discriminator co-trained to separate the policy's rollouts from black-box teacher responses, and many OPD methods such as GKD~\citep{agarwal2024policy} require teacher logits or token probabilities, which a black-box teacher does not expose.

\paragraph{Contribution of each expert and sensitivity to $\alpha$.} Table~\ref{tab:alpha} varies the expert weight $\alpha$ in Eq.~\eqref{eq:moe_reward}, where a larger $\alpha$ increases the weight of the perception expert; the endpoints correspond to perception-only ($\alpha{=}1$) and reasoning-only ($\alpha{=}0$) feedback. Each single expert improves over SFT$\rightarrow$GRPO (+2.0 and +3.0 points, respectively), and combining them reaches 66.2, exceeding the stronger single expert by 1.4 points. Perception-heavy settings are relatively stronger on general multimodal benchmarks, while reasoning-heavy settings perform better on most mathematical benchmarks. Intermediate weights remain stable (65.4--65.5 for $\alpha \in \{0.25, 0.75\}$), suggesting that the gain comes from complementary feedback rather than narrow tuning of $\alpha$.

\begin{table}[!htbp]
\caption{Sensitivity to the expert weight $\alpha$ on Qwen3-VL-4B (GRPO). $\alpha{=}0.5$ is the default PRISM setting. \textbf{Bold} indicates the best result in each column.}
\label{tab:alpha}
\centering
\small
\resizebox{\linewidth}{!}{
\begin{tabular}{c l cccc ccc c}
\toprule
$\alpha$ & \textbf{Setting} & \textbf{MathVista} & \textbf{MathVerse} & \textbf{MathVision} & \textbf{WeMath} & \textbf{MMMU} & \textbf{MMMU-Pro} & \textbf{HallusionBench} & \textbf{Avg.} \\
\midrule
-- & SFT$\rightarrow$GRPO & 75.7 & 64.5 & 35.5 & 77.8 & 60.1 & 47.3 & 72.0 & 61.8 \\
\midrule
1.00 & Perception only & 75.1 & 64.3 & 39.5 & 80.4 & 63.6 & 49.3 & 74.5 & 63.8 \\
0.75 & Perception + Reasoning & 77.6 & 67.7 & 44.0 & 81.8 & 63.8 & 49.0 & 74.7 & 65.5 \\
0.50 & Perception + Reasoning & \textbf{77.9} & \textbf{68.6} & \textbf{45.4} & \textbf{82.9} & \textbf{64.1} & \textbf{49.7} & \textbf{74.8} & \textbf{66.2} \\
0.25 & Perception + Reasoning & 77.3 & 68.1 & 43.8 & 82.4 & 62.9 & 49.3 & 73.9 & 65.4 \\
0.00 & Reasoning only & 76.9 & 67.4 & 43.6 & 81.8 & 62.4 & 48.4 & 73.3 & 64.8 \\
\bottomrule
\end{tabular}
}
\end{table}

\section{Analysis of the Post-SFT Distributional Mismatch}
\label{app:drift}

\paragraph{What distribution is aligned.} Throughout the paper, \emph{distributional drift} refers to the change in the policy-induced trajectory distribution $\pi_\theta(\tau \mid x)$, where a trajectory $\tau = (c, t, a)$ consists of the visual description $c$, the reasoning trace $t$, and the final answer $a$. We do not estimate a formal divergence between trajectory distributions, which is intractable in this high-dimensional output space. Instead, we report two complementary diagnostics: within-policy rollout statistics, which characterize RL-relevant properties of each checkpoint's own distribution, and a teacher-referenced embedding similarity, which measures semantic proximity to the supervision responses. Neither is a direct measure of distributional distance; together with the structural statistics in Section~\ref{sec:analysis}, they characterize different aspects of the observed behavior.

\paragraph{Rollout statistics.} We hypothesize that SFT reduces rollout diversity and the within-group correctness variation that group-based RL relies on. We sample 16 rollouts per prompt from each pre-RL checkpoint under identical settings and report the answer entropy, the average within-prompt standard deviation of binary correctness rewards, the informative-group ratio (the percentage of prompts whose rollouts contain both correct and incorrect answers), and the pass rate (Table~\ref{tab:rollout_stats}). Compared with the Instruct model, SFT lowers answer entropy, reward variation, and the informative-group ratio, leaving fewer useful within-group comparisons for RLVR. Pre-alignment reverses this trend: relative to the post-SFT checkpoint, it increases answer entropy by 12.1\%, the reward standard deviation by 22.3\%, and the informative-group ratio from 64.1\% to 75.4\%, while the pass rate changes only slightly (39.7\% to 40.6\%). Thus, pre-alignment mainly improves the diversity and discriminative quality of rollout groups rather than directly increasing correctness, providing a more learnable initialization for RLVR.

\begin{table}[!htbp]
\caption{Rollout statistics of pre-RL checkpoints on Qwen3-VL-4B (16 rollouts per prompt).}
\label{tab:rollout_stats}
\centering
\small
\begin{tabular}{l cccc}
\toprule
\textbf{Checkpoint} & \textbf{Answer entropy} & \textbf{Avg.\ reward std.} & \makecell{\textbf{Informative-}\\\textbf{group ratio}} & \textbf{Pass rate} \\
\midrule
Instruct & 0.467 & 0.244 & 70.8\% & 44.3\% \\
Post-SFT & 0.438 & 0.211 & 64.1\% & 39.7\% \\
Post-alignment & 0.491 & 0.258 & 75.4\% & 40.6\% \\
\bottomrule
\end{tabular}
\end{table}

\paragraph{Teacher-referenced embedding similarity.} Caption-item and reasoning-step counts are coarse structural statistics. As semantic evidence beyond these counts, we take 2,000 prompts, remove the final-answer span from each response, embed the remaining perception and reasoning content with the frozen Qwen3-Embedding-8B model~\citep{zhang2025qwen3}, and compute the paired cosine similarity to the corresponding Gemini response (Table~\ref{tab:embed_sim}). SFT substantially increases similarity to Gemini, and pre-alignment further increases it from 0.851 to 0.872. The similarity remains high after RLVR, with a small decrease, plausibly because RLVR optimizes correctness rather than imitation of the supervision responses. Notably, the post-SFT checkpoint is already semantically close to Gemini, yet GRPO from it reaches only 61.8 compared with 66.2 from the post-alignment checkpoint. Semantic proximity is therefore beneficial but insufficient to explain the gain on its own, consistent with the rollout statistics above showing that pre-alignment also improves the learning signals available to RLVR. We note that this measure is teacher-referenced and depends on the choice of embedding model.

\begin{table}[!htbp]
\caption{Paired cosine similarity between each checkpoint's perception and reasoning content and the corresponding Gemini response (Qwen3-VL-4B, 2,000 prompts, Qwen3-Embedding-8B).}
\label{tab:embed_sim}
\centering
\small
\begin{tabular}{lc}
\toprule
\textbf{Checkpoint} & \textbf{Similarity to Gemini} \\
\midrule
Instruct & 0.778 \\
Post-SFT & 0.851 \\
Post-alignment (PRISM) & 0.872 \\
Post-alignment + GRPO & 0.868 \\
\bottomrule
\end{tabular}
\end{table}

\section{Data Curation Pipeline}
\label{app:data}

We describe the full pipeline for constructing our curated 113K multimodal reasoning dataset. The process consists of source preparation, iterative generation with multi-stage filtering, and downstream data splitting.

\paragraph{Source preparation.}
We begin with a collection of multimodal reasoning problems sourced from public multimodal reasoning datasets, including LLaVA-CoT~\citep{xu2025llava}, OpenVLThinker~\citep{deng2026openvlthinker}, and We-Math~2.0~\citep{qiao2025wemath2}, spanning mathematical reasoning, scientific diagram understanding, chart interpretation, and spatial reasoning. During initial inspection, we observe that a non-trivial fraction of problems contain multiple sub-questions bundled under a single image but are paired with only one ground-truth answer. To resolve this ambiguity, we concatenate the full problem and its ground-truth answer, then prompt an LMM to identify which sub-question the answer corresponds to, discarding the remaining sub-questions. This ensures a clean one-to-one mapping between each problem and its ground truth before generation begins.

\paragraph{Distillation and iterative filtering.}
We prompt Gemini 3 Flash with a carefully designed distillation template (see Appendix~\ref{app:examples}) that requires fine-grained visual descriptions, step-by-step reasoning traces, and concise final answers. The raw generations are then processed through a three-stage filtering pipeline:
\begin{enumerate}[nosep,leftmargin=*]
    \item \textbf{Truncation and failure filter:} We remove responses that are truncated due to exceeding the maximum output length, as well as those where the API call fails.
    \item \textbf{Format filter:} We verify that each response contains the required output structure (\texttt{<caption>}, \texttt{<think>}, \texttt{<answer>} tags) with non-empty content in each section.
    \item \textbf{Correctness filter:} We apply an LLM-as-judge to compare the generated answer against the ground truth, discarding responses whose answers are deemed incorrect.
\end{enumerate}
To maximize data yield, samples that fail the truncation/failure filter or the format filter are re-submitted to Gemini 3 Flash for regeneration. Samples that fail the correctness filter are re-submitted with the ground-truth answer appended to the prompt, guiding the model toward a correct solution while preserving the requirement for detailed reasoning. After regeneration, all re-submitted samples pass through the same three-stage filter again. This iterative process substantially improves the overall yield without sacrificing quality.

\paragraph{Data splitting.}
From the resulting 113K verified samples, we randomly hold out 6K samples for the alignment and RL stages, and assign the remaining 107K to the SFT pool. For the RLVR training set, we further apply a difficulty-based filter: we sample $N=16$ rollouts per problem from the post-alignment policy and retain only problems whose pass rate falls within $[0.2, 0.8]$, yielding a final set of approximately 2K difficulty-matched samples. This ensures that RLVR trains on problems that are neither trivially easy nor prohibitively hard for the current policy. The same 2K subset is used by all RLVR runs, including the baselines. Table~\ref{tab:data_summary} summarizes the data composition across all stages.

\paragraph{Deduplication and contamination checks.} We deduplicate the curated corpus at the original question--image level, yielding 113,801 examples, which are split into 107,801 SFT examples and a disjoint 6K pool. The SFT split is combined with the 1,268,644 public MMR1 examples~\citep{leng2025mmr1}. The 2K RLVR subset is drawn from the 6K pool, so the overlap between the alignment and RLVR prompts is intentional. We checked the complete SFT corpus, the 6K pool, and the 2K RLVR subset against all seven evaluation benchmarks and found no exact question--image overlap; no evaluation sample is used in SFT, pre-alignment, or RLVR.

\begin{table}[h]
\caption{Summary of data used across PRISM stages.}
\label{tab:data_summary}
\centering
\small
\begin{tabular}{llr}
\toprule
\textbf{Stage} & \textbf{Source} & \textbf{Samples} \\
\midrule
SFT & Gemini 3 Flash (curated) & 107K \\
SFT & Public demonstrations~\citep{leng2025mmr1} & 1.26M \\
\midrule
Alignment & Gemini 3 Flash (curated) & 6K \\
\midrule
RLVR & Difficulty-filtered subset & 2K \\
\bottomrule
\end{tabular}
\end{table}

\section{Examples}
\label{app:examples}
\subsection{Prompts and Data Examples}
\begin{figure}[h]
\centering
\includegraphics[width=1.0\linewidth]{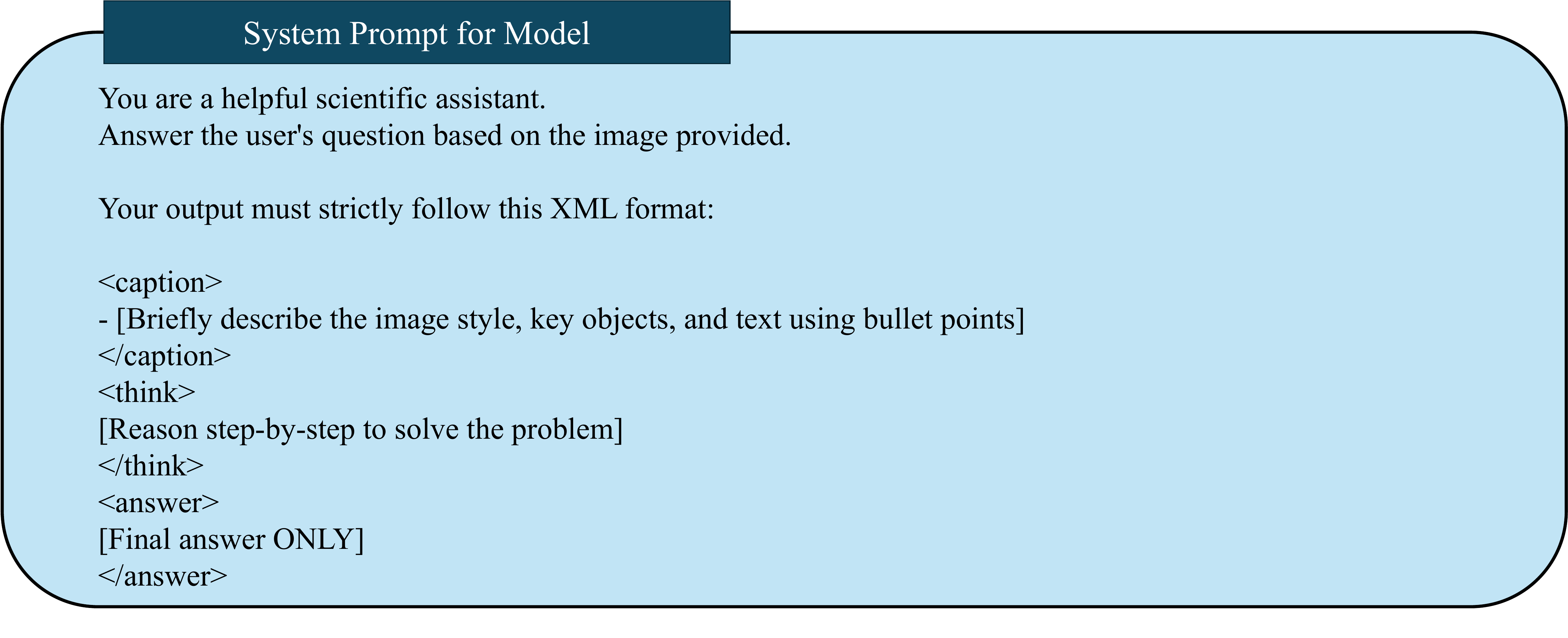}
\caption{\textbf{System Prompt.} This system prompt is shared across SFT, RL training, and benchmark evaluation to enforce the structured three-part output format (\texttt{<caption>}, \texttt{<think>}, \texttt{<answer>}).}
\label{fig:prompt-training}
\end{figure}

\begin{figure}[htbp]
\centering
\includegraphics[width=1.0\linewidth]{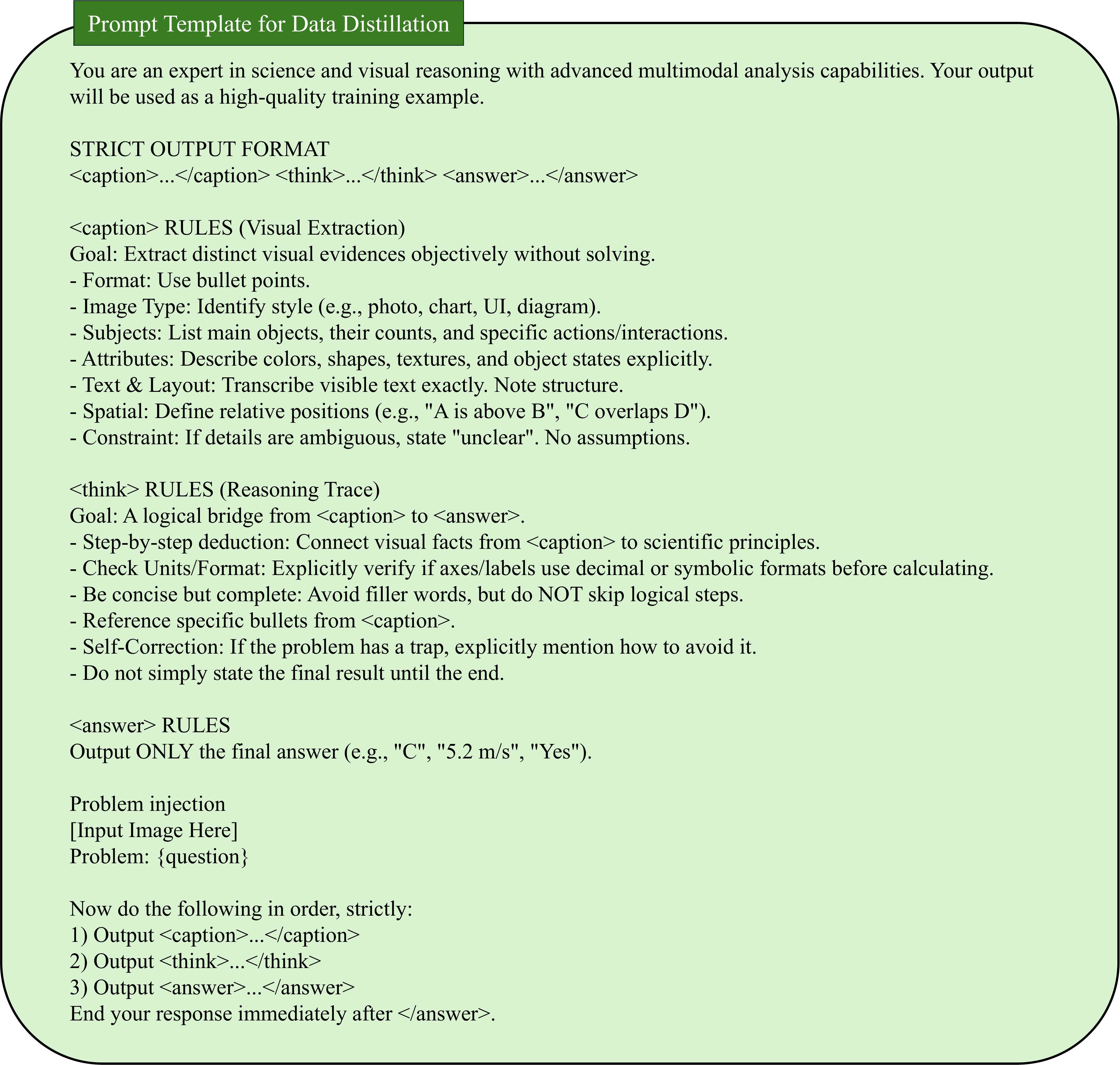}
\caption{\textbf{Data Distillation Prompt.} The full prompt used to query Gemini 3 Flash for generating high-quality multimodal reasoning demonstrations with explicit rules for visual extraction, reasoning traces, and concise answers.}
\label{fig:prompt-distillation}
\end{figure}

\begin{figure}[htbp]
\centering
\includegraphics[width=1.0\linewidth]{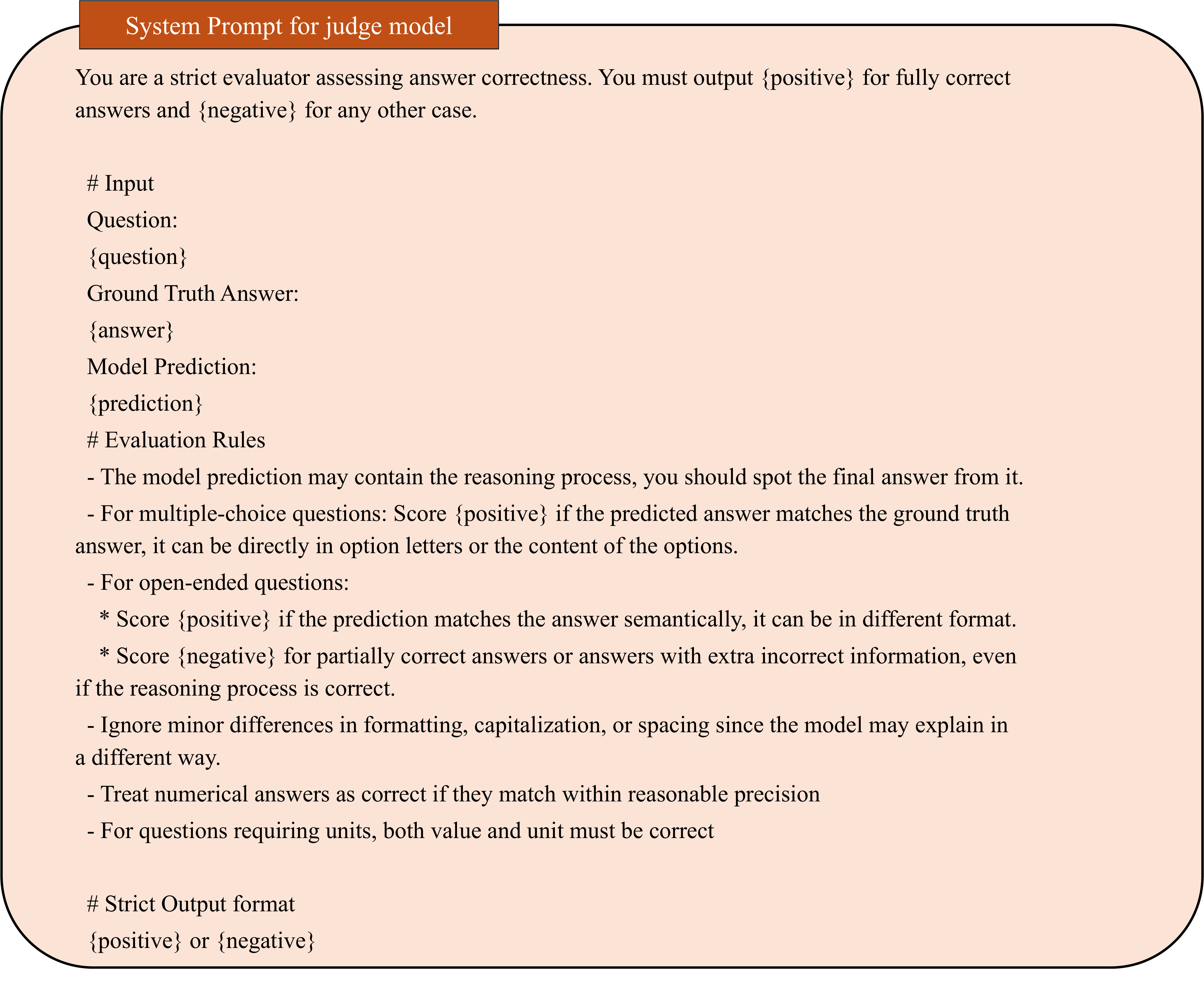}
\caption{\textbf{Judge Model Prompt.} The prompt used for LLM-as-judge evaluation, where Qwen3-30B-A3B-Instruct compares the model's extracted answer against the ground truth.}
\label{fig:prompt-judge}
\end{figure}

We provide the prompts used across different stages of PRISM. Figure~\ref{fig:prompt-training} shows the system prompt shared across SFT, RL training, and evaluation, which enforces a structured three-part output format: \texttt{<caption>} for visual grounding, \texttt{<think>} for step-by-step reasoning, and \texttt{<answer>} for the final response. Figure~\ref{fig:prompt-distillation} presents the distillation prompt used to elicit detailed reasoning demonstrations from Gemini 3 Flash, which includes fine-grained rules for each output section to ensure high-quality supervision. Figure~\ref{fig:prompt-judge} shows the prompt used for LLM-as-judge evaluation.

\subsection{Cold-Start Data Samples}
\begin{figure}[htbp]
    \centering
    \includegraphics[width=1.0\linewidth]{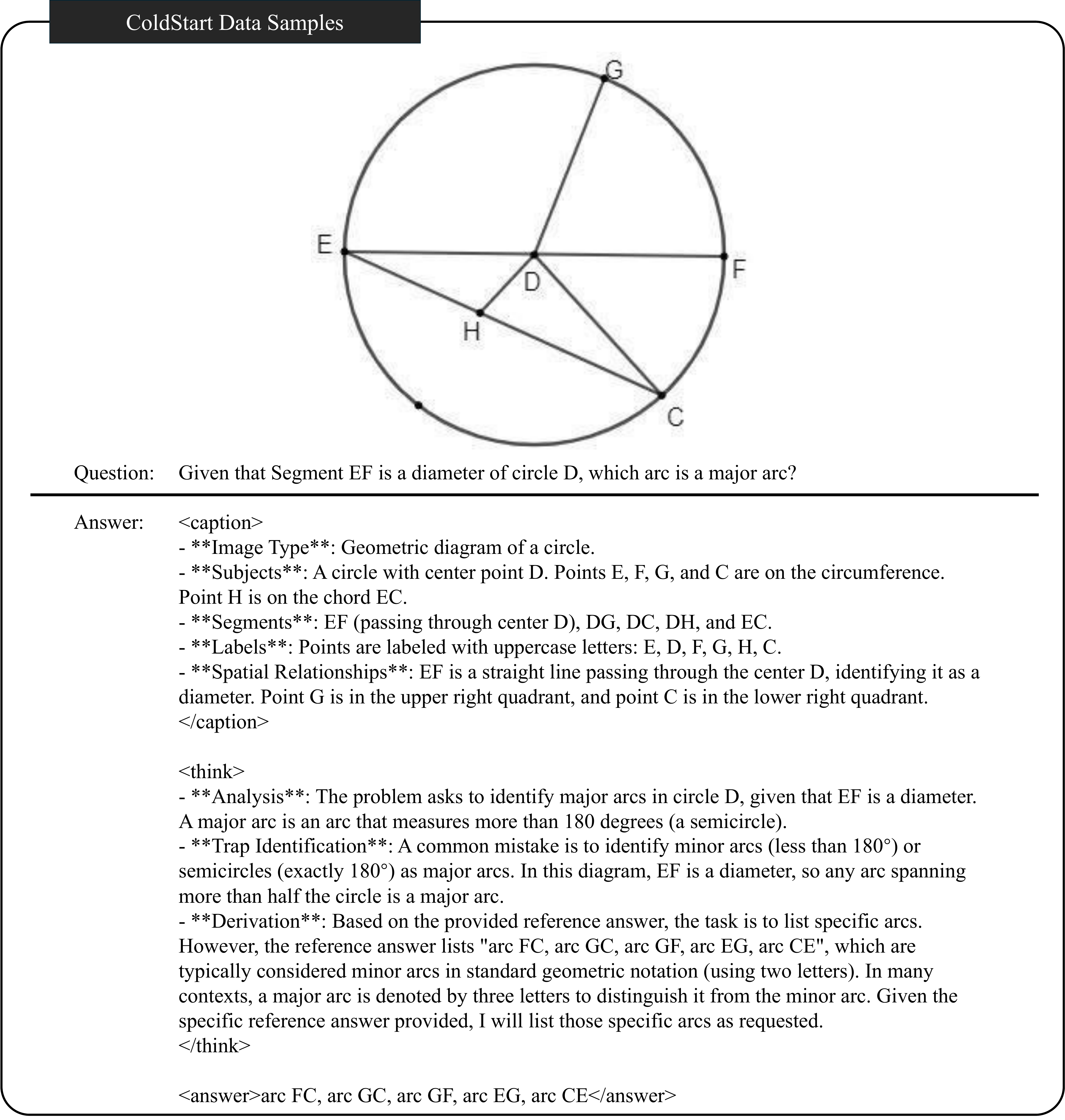}
    \caption{An example of a cold-start data sample.}
    \label{fig:sft-example1}
\end{figure}

\begin{figure}[htbp]
    \centering
    \includegraphics[width=1.0\linewidth]{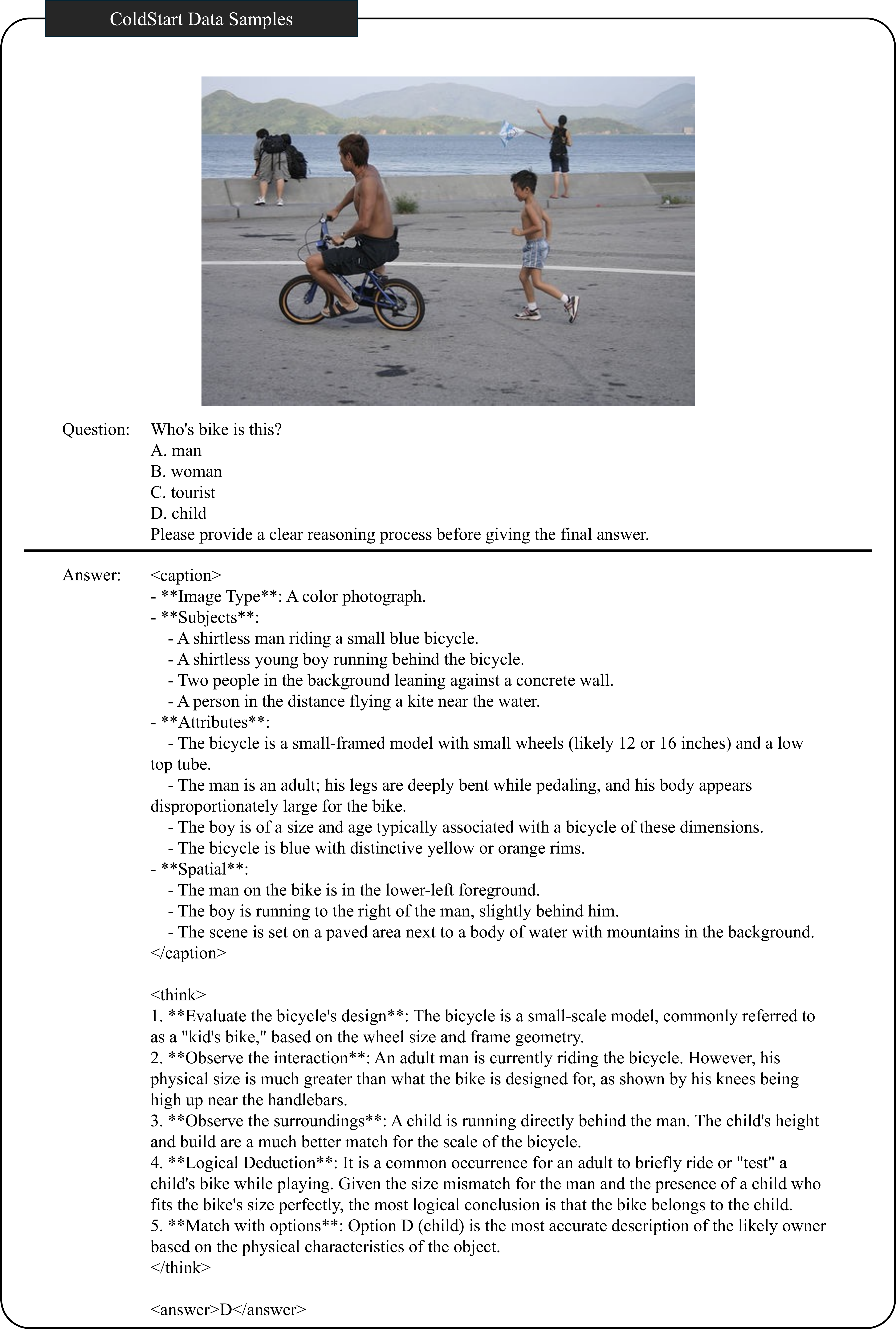}
    \caption{An example of a cold-start data sample.}
    \label{fig:sft-example2}
\end{figure}

We present representative examples from the cold-start SFT dataset used in Stage~1 of PRISM. Figure~\ref{fig:sft-example1} and Figure~\ref{fig:sft-example2} illustrate the structured three-part output format, including visual grounding via \texttt{<caption>}, chain-of-thought reasoning via \texttt{<think>}, and the final response via \texttt{<answer>}.

\subsection{Qualitative Data Samples}
\begin{figure}[htbp]
    \centering
    \includegraphics[width=1.0\linewidth, height=0.98\textheight, keepaspectratio]{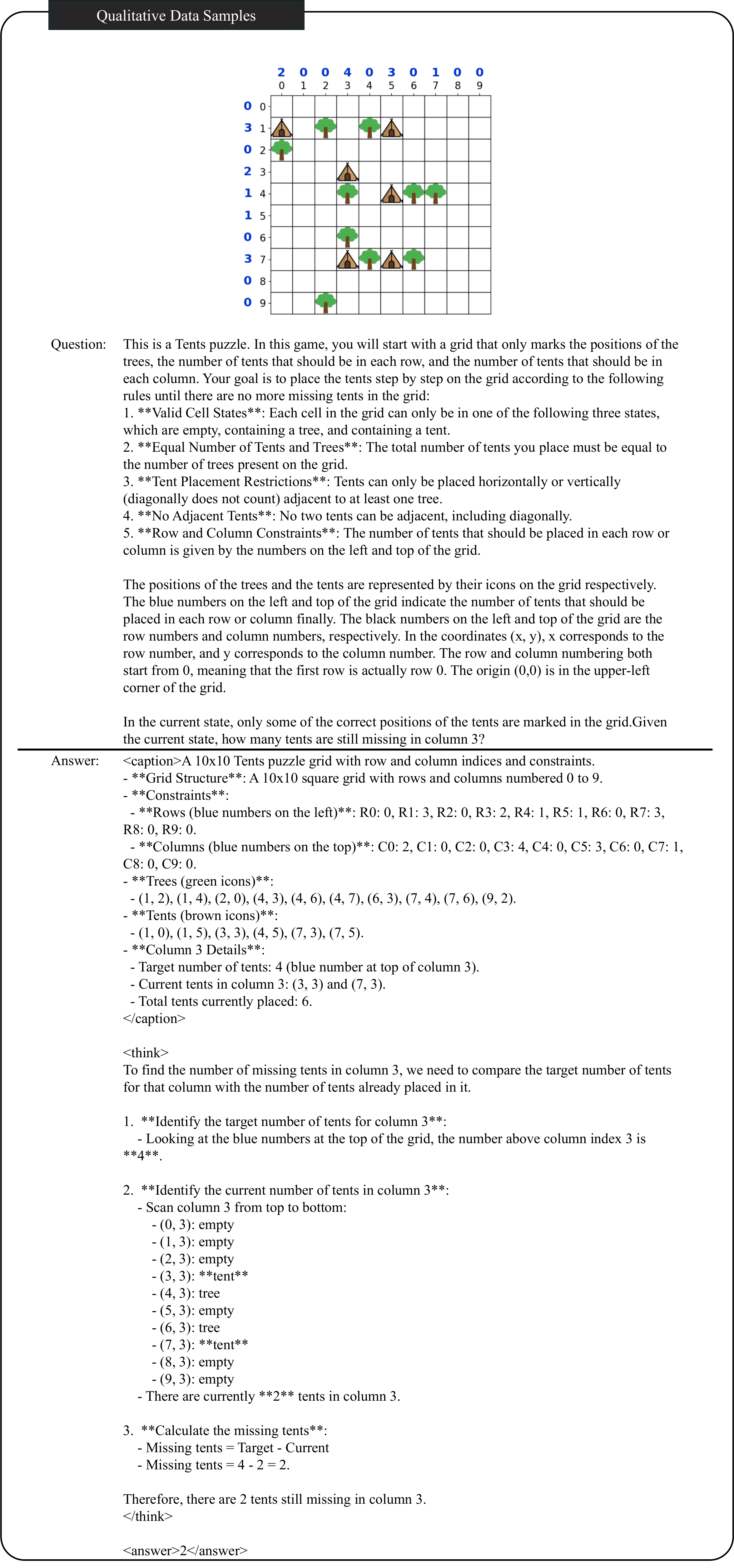}
    \caption{An example of our model inference result.}
    \label{fig:rollout-example1}
\end{figure}

\begin{figure}[htbp]
    \centering
    \includegraphics[width=1.0\linewidth, height=0.98\textheight, keepaspectratio]{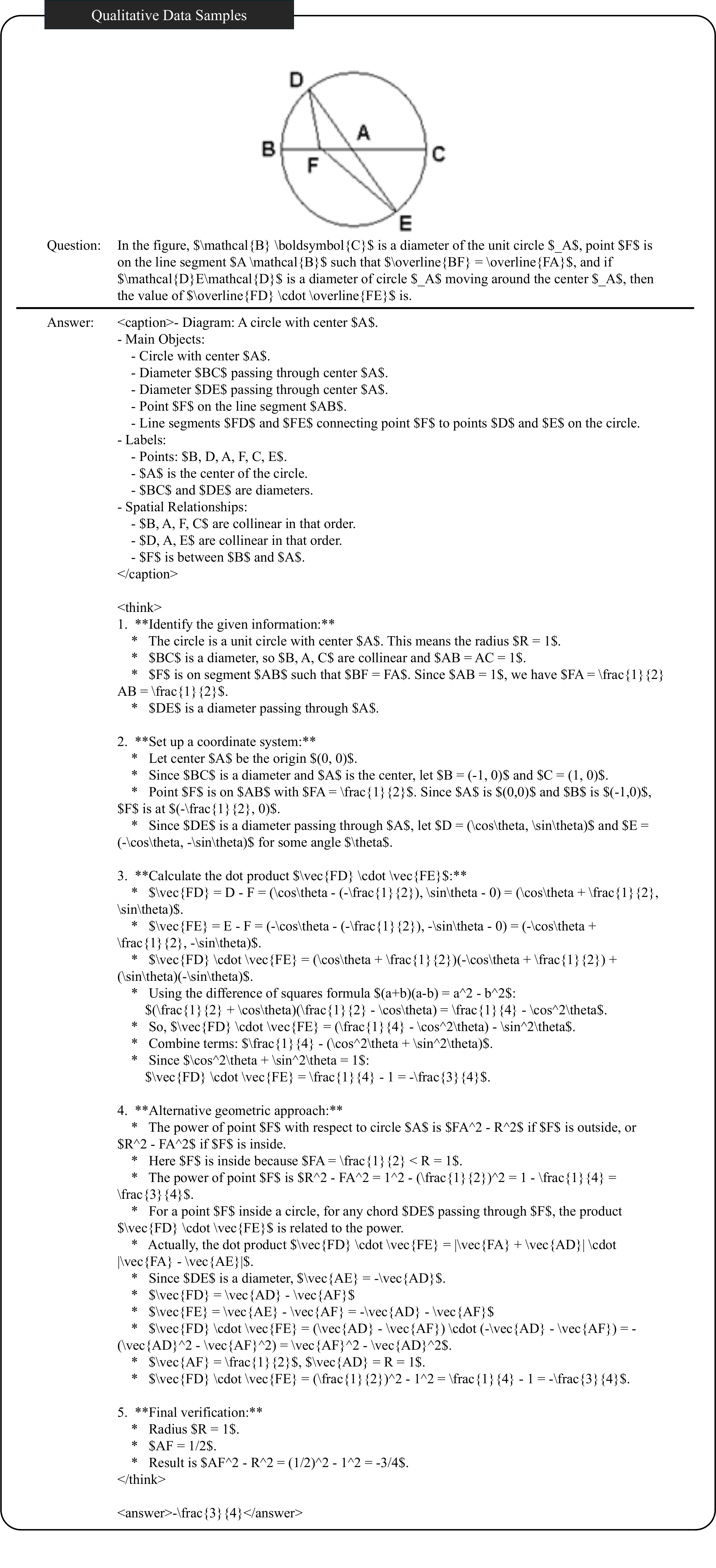}
    \caption{An example of our model inference result.}
    \label{fig:rollout-example2}
\end{figure}

We provide qualitative examples of model rollouts generated during the reinforcement learning stage after PRISM. Figure~\ref{fig:rollout-example1} and Figure~\ref{fig:rollout-example2} show representative reasoning trajectories produced by the policy model, demonstrating the structured caption-think-answer pipeline in practice.

\clearpage
\section{Full Training Procedure}
\label{app:algorithm}

We provide the complete PRISM training procedure in Algorithm~\ref{alg:prism}. The pipeline consists of three sequential stages. Stage~1 performs standard SFT on the combined corpus to obtain an initial policy $\pi_{\text{sft}}$. Stage~2 alternates between updating the MoE discriminator (via Bradley-Terry loss on supervision vs.\ policy rollouts) and updating the policy (via GRPO with discriminator rewards), driving the policy distribution toward the supervision distribution. Stage~3 switches the reward from the discriminator to a deterministic verifiable reward and runs outcome-based RLVR on the difficulty-filtered subset. Each stage feeds its output checkpoint as the initialization for the next.

\begin{algorithm}[htbp]
\caption{PRISM: Three-Stage Post-Training Pipeline}
\label{alg:prism}
\begin{algorithmic}[1]
\REQUIRE SFT corpus $\mathcal{D}_{\text{SFT}}$, curated supervision data $\mathcal{T}_{\text{align}}$, base model $\pi_0$, balancing weight $\alpha$, group size $N$
\ENSURE Trained policy $\pi^*$

\STATE \textbf{// Stage 1: SFT Cold Start}
\STATE $\pi_{\text{sft}} \leftarrow \text{SFT}(\pi_0, \mathcal{D}_{\text{SFT}})$

\STATE \textbf{// Stage 2: Distribution Alignment via On-Policy Distillation}
\STATE Initialize policy $G \leftarrow \pi_{\text{sft}}$
\STATE Initialize perception expert $D_v$ and reasoning expert $D_r$ from pretrained backbone; warm-start on $\mathcal{T}_{\text{align}}$
\FOR{each training step}
    \STATE Sample prompts $\{x\}$ from $\mathcal{T}_{\text{align}}$
    \STATE Generate rollouts $y^{-} \sim G(\cdot | x)$; extract $c^{-}, t^{-}$
    \STATE Retrieve supervision responses $y^{+}$ from $\mathcal{T}_{\text{align}}$; extract $c^{+}, t^{+}$
    \STATE \textbf{Update discriminator:}
    \STATE \quad $\mathcal{L}_{D_v} \leftarrow -\log \sigma(D_v(x, c^{+}) - D_v(x, c^{-}))$ \hfill $\triangleright$ Perception expert
    \STATE \quad $\mathcal{L}_{D_r} \leftarrow -\log \sigma(D_r(x, t^{+}) - D_r(x, t^{-}))$ \hfill $\triangleright$ Reasoning expert
    \STATE \quad Update $D_v, D_r$ with $\mathcal{L}_{D_v} + \mathcal{L}_{D_r}$
    \STATE \textbf{Update policy:}
    \STATE \quad Sample $N$ responses $\{y^{-}_i\}_{i=1}^N$ per prompt from $G$
    \STATE \quad $r_i \leftarrow \alpha \cdot D_v(x, c^{-}_i) + (1 - \alpha) \cdot D_r(x, t^{-}_i)$ \hfill $\triangleright$ MoE reward
    \STATE \quad Compute advantages $A_i$ via group normalization
    \STATE \quad Update $G$ via GRPO with $\{A_i\}$
\ENDFOR
\STATE $\pi_{\text{align}} \leftarrow G$

\STATE \textbf{// Stage 3: RLVR}
\STATE Initialize policy $\pi \leftarrow \pi_{\text{align}}$
\FOR{each training step}
    \STATE Sample prompts, generate rollouts, compute $r_{\text{v}} = 0.8\, r_{\text{acc}} + 0.2\, r_{\text{fmt}}$
    \STATE Update $\pi$ via GRPO/DAPO/GSPO with verifiable rewards
\ENDFOR
\STATE \RETURN $\pi^* \leftarrow \pi$
\end{algorithmic}
\end{algorithm}

\section{Limitations and Future Directions}
\label{app:limitations}

PRISM introduces additional training cost due to the alignment stage and the MoE discriminator. Although the alignment stage runs for only 500 steps, it requires jointly maintaining the policy generator and the discriminator during training, increasing memory and compute overhead compared to a standard SFT$\rightarrow$RLVR pipeline; our comparisons are therefore matched in policy optimization updates rather than FLOPs. Our experiments are also limited to a single model family (Qwen3-VL), the additional controlled comparisons in Appendix~\ref{app:controls} are conducted only on 4B, and the discriminator is built from the same model family as the policy. In addition, the current MoE design relies on structured response formats (explicit visual caption and reasoning trace) to decompose perception and reasoning feedback, which may limit direct applicability to tasks that do not naturally admit such decomposition. Moreover, our evidence for distribution alignment (structural proxies in Section~\ref{sec:analysis}; rollout statistics and teacher-referenced embedding similarity in Appendix~\ref{app:drift}) does not constitute a formal measure of distributional distance; developing model-agnostic alignment metrics remains an open problem. Future work includes validating PRISM on other model families, extending PRISM to broader task formats by exploring automatic or learned decomposition of discriminator feedback, scaling to larger base models, and investigating whether the alignment stage can be further shortened or amortized across multiple downstream RL objectives. More broadly, by improving the faithfulness of multimodal reasoning, PRISM may contribute to more reliable vision-language systems; however, as with all advances in large model training, downstream applications should be evaluated for potential misuse.

\end{document}